\documentclass{article}

\usepackage{iclr2027_conference,times}
\iclrfinalcopy

\usepackage[T1]{fontenc}
\usepackage[hidelinks]{hyperref}
\usepackage{url}
\usepackage{booktabs}
\usepackage{amsmath,amssymb,amsfonts}
\usepackage{nicefrac}
\usepackage{microtype}
\usepackage{xcolor}
\usepackage{graphicx}
\usepackage{wrapfig}
\usepackage{array}
\usepackage{multirow}
\usepackage{colortbl}
\usepackage[most]{tcolorbox}
\tcbuselibrary{breakable,skins}
\usepackage{algorithm}   
\usepackage{xspace}
\usepackage{needspace}
\usepackage{etoolbox}
\makeatletter
\let\originalsection\section
\renewcommand{\section}{\@ifstar{\originalsection*}{\layoutsection}}
\newcommand{\layoutsection}[1]{%
  \ifstrequal{#1}{Related Work}{\Needspace{10\baselineskip}}{}%
  \ifstrequal{#1}{Extended Related Work}{\Needspace{10\baselineskip}}{}%
  \originalsection{#1}}
\makeatother
\let\originalsubsection\subsection
\renewcommand{\subsection}[1]{%
  \ifstrequal{#1}{Skill Design, Acquisition, and Refinement}{\Needspace{6\baselineskip}}{}%
  \originalsubsection{#1}}

\newcommand{\Name}{\texttt{SkillFusion}\xspace}
\newcommand{\contribone}{\textbf{(1)}\xspace}
\newcommand{\contribtwo}{\textbf{(2)}\xspace}
\newcommand{\contribthree}{\textbf{(3)}\xspace}

\definecolor{tab-row-even}{HTML}{F4F7FA}
\definecolor{tab-highlight}{HTML}{E8EEF5}
\definecolor{sysbox-bg}{HTML}{EBF5FB}
\definecolor{sysbox-frame}{HTML}{2980B9}
\definecolor{userbox-bg}{HTML}{FEF9E7}
\definecolor{userbox-frame}{HTML}{D4AC0D}
\definecolor{advbox-bg}{HTML}{FDEDEC}
\definecolor{advbox-frame}{HTML}{C0392B}
\definecolor{notebox-bg}{HTML}{E8F8F5}
\definecolor{notebox-frame}{HTML}{1ABC9C}

\newtheorem{definition}{Definition}

\newtcolorbox{systempromptbox}[1][]{%
  breakable, lines before break=4,
  colback=sysbox-bg, colframe=sysbox-frame,
  boxrule=0.6pt, arc=2.5pt,
  coltitle=white, fonttitle=\small\bfseries\sffamily,
  title={\raisebox{-0.1ex}{\small$\blacktriangleright$}\;\;System Prompt},
  top=5pt, bottom=4pt, left=8pt, right=8pt,
  before skip=4pt plus 1pt, after skip=4pt plus 1pt,
  #1
}

\newtcolorbox{userpromptbox}[1][]{%
  breakable, lines before break=4,
  colback=userbox-bg, colframe=userbox-frame,
  boxrule=0.6pt, arc=2.5pt,
  coltitle=white, fonttitle=\small\bfseries\sffamily,
  title={\raisebox{-0.1ex}{\small$\blacktriangleright$}\;\;User Prompt},
  top=5pt, bottom=4pt, left=8pt, right=8pt,
  before skip=4pt plus 1pt, after skip=4pt plus 1pt,
  #1
}

\newtcolorbox{adversarialpromptbox}[1][]{%
  breakable, lines before break=4,
  colback=advbox-bg, colframe=advbox-frame,
  boxrule=0.6pt, arc=2.5pt,
  coltitle=white, fonttitle=\small\bfseries\sffamily,
  title={\raisebox{-0.1ex}{\small$\blacktriangleright$}\;\;Adversarial Second-Pass},
  top=5pt, bottom=4pt, left=8pt, right=8pt,
  before skip=4pt plus 1pt, after skip=4pt plus 1pt,
  #1
}

\newtcolorbox{notebox}[1][]{%
  breakable, lines before break=4,
  colback=notebox-bg, colframe=notebox-frame,
  boxrule=0.5pt, arc=2.5pt,
  coltitle=white, fonttitle=\small\bfseries\sffamily,
  title={\raisebox{-0.1ex}{\small$\bigstar$}\;\;Implementation Note},
  top=5pt, bottom=4pt, left=8pt, right=8pt,
  before skip=4pt plus 1pt, after skip=4pt plus 1pt,
  #1
}

\title{\Name: Towards Reliable Multi-Skill Composition for LLM Agents}

\author{Enhao Huang$^{1}$, Xinye Zhao$^{2}$, Yiming Li$^{3}$\thanks{Corresponding author.}, Bingrun Yang$^{1}$,\\
\bfseries Nanjun Zhou$^{1}$, Chunshu Xia$^{1}$, Hongwei Yao$^{4}$, Zhan Qin$^{1}$\\[3pt]
\normalfont $^{1}$Zhejiang University \quad $^{2}$Nanjing University\\
\normalfont $^{3}$Nanyang Technological University\\
\normalfont $^{4}$City University of Hong Kong}

\renewcommand{\topfraction}{0.90}
\renewcommand{\bottomfraction}{0.85}
\renewcommand{\textfraction}{0.08}
\renewcommand{\floatpagefraction}{0.80}
\makeatletter
\patchcmd{\@maketitle}
  {{\LARGE\sc \@title\par}}
  {\toptitlebar{\LARGE\sc \@title\par}\bottomtitlebar}
  {}{\PackageError{arxiv-layout}{Could not add ICLR title bars}{}}
\makeatother
\begin{document}
\maketitle
\fancyhead{}
\renewcommand{\headrulewidth}{0pt}

\begin{abstract}
Independently authored skills can impose incompatible requirements, competing preferences, and redundant instructions when jointly loaded by an LLM agent. We propose \Name, an offline compiler that reconciles these interactions into a reusable merged skill. Its central idea is to make composition decisions explicit through typed, scoped constraints linked to their source directives. A type--scope gate selects candidate interactions, while selective fallback revisits pairs potentially missed by structural pruning. A classifier identifies interaction types, and an adversarial verifier re-examines uncertain judgments. The resolver applies conditional routing or explicit priorities to hard conflicts, verified abstraction to soft conflicts, and deduplication to redundant instructions. Source-linked checks guide synthesis and compression, and metadata synthesis supports discovery of the merged skill. The resulting artifact retains associated scripts and resources and can be reused across tasks and executor models. \Name achieves the highest mean Success among evaluated non-oracle methods on all three agent benchmarks and on both transfer benchmarks across three executor models. At the largest tested instruction configurations, it achieves over $2.5\times$ the Success of unresolved multi-skill loading, highlighting the growing value of reconciliation as composition becomes more demanding.
The code will be open-sourced after the review process concludes.
\end{abstract}

\section{Introduction}
\label{sec:intro}

LLM agents use reusable \emph{skills} to carry domain knowledge into tasks such as spreadsheet editing, browser interaction, and software maintenance. A skill packages instructions, scripts, and resources around a \texttt{SKILL.md} file; metadata lets the agent discover and load the package when needed~\citep{anthropic2026skillsguide, openai2026codex}. This modular design lets authors encode workflows once and reuse them across tasks. As tasks span several workflows, an agent may activate skills from multiple authors in the same execution context~\citep{openai2026skillsblog}. Their joint use creates a composition problem: the agent must determine how independently written instructions should jointly govern a task.

\begin{figure}[t]
    \centering
    
    \includegraphics[width=0.98\linewidth]{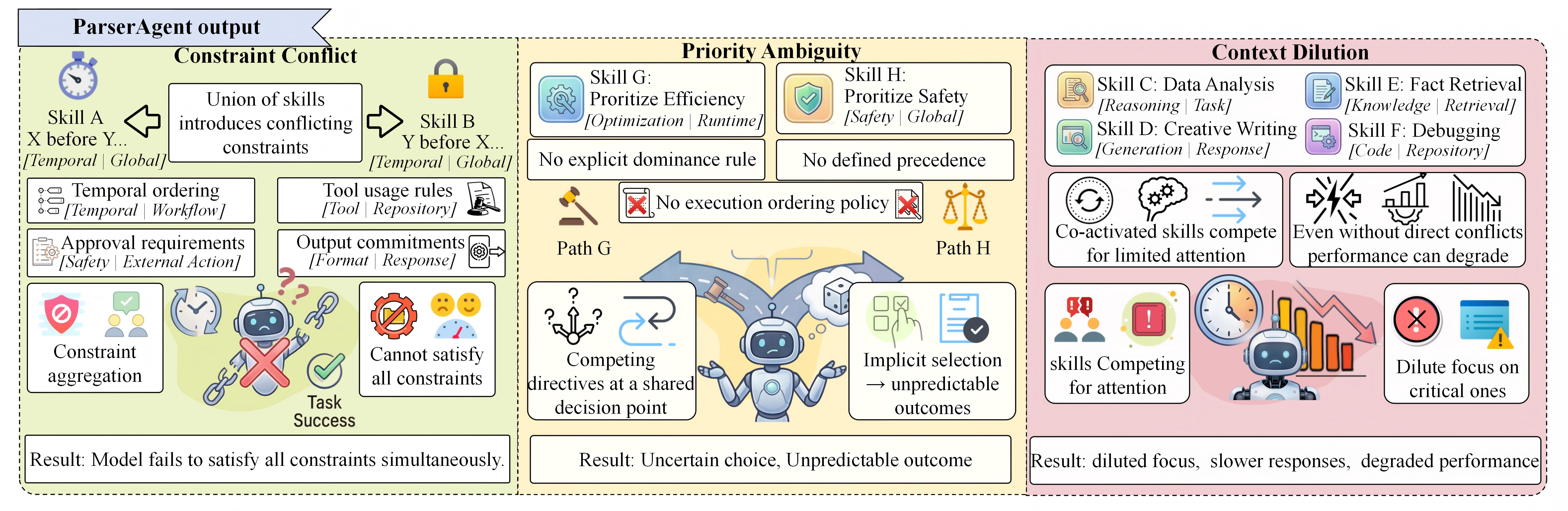}
    
    \caption{Illustrative examples of three failure modes in multi-skill systems: \emph{hard conflict}, \emph{soft conflict}, and \emph{redundancy}. These arise when independently authored skills are co-activated, leading to incompatible constraints, competing directives, and redundant instructions, respectively.}
    \label{fig:multi-skill-challenges}
    
\end{figure}

The difficulty lies in the interactions between otherwise useful directives. Consider two spreadsheet skills: one recommends \texttt{openpyxl} for editing formulas and formatting, while another prescribes standard-library-only workbook auditing. Their applicability depends on the workflow phase. Removing that distinction can turn a useful division of responsibilities into an apparent conflict. More generally, co-activated skills introduce three interaction types (Figure~\ref{fig:multi-skill-challenges}). \emph{Hard conflicts} demand incompatible actions at the same decision point. \emph{Soft conflicts} introduce competing priorities, such as concise code and extensive explanatory comments. \emph{Redundancy} repeats overlapping requirements, increasing the instruction burden. Reliable composition therefore requires both a decision about how directives interact and an account of when each directive applies.

Existing work develops complementary parts of the skill ecosystem. Skill acquisition and refinement construct reusable knowledge~\citep{wang2026skillx}, while retrieval and routing select relevant skills from large, noisy libraries~\citep{zheng2026skillrouter, zhang2026benchmarking}. Public-library analyses also report substantial intent-level redundancy~\citep{ling2026agentskills}. Once several skills have been selected, concatenation leaves their reconciliation to the executor. Free-form rewriting can consolidate their prose, but its treatment of individual requirements and exceptions is difficult to inspect. Instruction-following studies reinforce this concern: composed constraints challenge LLMs~\citep{wen2024complexbench}, accumulating requirements reduces performance~\citep{guo2025recast}, and conflicting instructions can lead to partial or silently compromised compliance~\citep{qian2025coninstruct}.

We ask: \emph{how can independently authored skills be merged into a coherent operational specification while retaining their behavioral requirements?} Our key insight is that the appropriate merge operation depends on the \emph{type and scope of the interaction}. Incompatible requirements need a routing or priority decision; competing preferences need a precise common formulation; duplicate requirements can be consolidated. Applying these operations requires an explicit representation of source directives and their activation conditions. It also requires selective reconsideration: an early error in typing or scope estimation can exclude a relevant interaction before the resolver sees it.

Based on this insight, we propose \Name, an offline compiler for instruction-level skill composition. A Parser Agent extracts typed, scoped constraints and links them to their sources. A type--scope gate identifies likely interactions, and selective fallback revisits excluded pairs with uncertain typing or scope, global scope, or high semantic similarity. A Classifier Agent labels candidate interactions, with adversarial re-judging for uncertain verdicts. The Resolver Agent then applies routing or priority decisions, verified abstraction, and deduplication according to the interaction type. Source-linked checks accompany synthesis and compression, and the output retains discovery metadata, scripts, and resources. The resulting skill is compiled once per pool and reused across tasks and executor models.

Experiments connect this design to both task performance and interaction recovery. On SpreadsheetBench Verified~\citep{ma2024spreadsheetbench}, WebArena-Verified~\citep{zhou2024webarena}, and SWE-bench Verified~\citep{jimenez2023swe,openai2024swebenchverified}, \Name improves Success over three-call Rewrite--Critique--Revise by 4.8, 4.0, and 3.8 percentage points, respectively. A human audit shows that selective fallback recovers 14 of the 24 incompatible pairs missed by strict pruning, increasing F1 from .713 to .840. Controlled mechanism experiments reveal performance drops under each interaction type and their joint condition, while component ablations assess the corresponding design choices. Transfer across three executor models and scaling to 140 constraints further evaluate the reuse and robustness of the compiled artifacts. The widening advantage over unresolved loading indicates that reconciliation becomes more valuable as instruction complexity grows, while cross-model transfer supports reuse beyond the compilation backbone.

Our contributions are threefold. \contribone We formulate instruction-level skill composition through typed, scoped constraints and distinguish hard conflicts, soft conflicts, and redundancy as targets for reconciliation. \contribtwo We develop selective candidate generation and interaction-specific resolution with source-linked verification, producing reusable merged skills. \contribthree We evaluate task performance on three agent benchmarks and connect the results to interaction recovery through human auditing, controlled mechanisms, component ablations, cross-model transfer, and instruction scaling.

\section{Related Work}
\label{sec:related}

We briefly position \Name relative to four lines of work; a comprehensive survey with per-paper discussion is provided in Appendix~\ref{app:extended_related}.

\paragraph{Skill construction and routing.}
A rapidly growing body of work addresses both how to build individual skills and how to reliably select the right one at inference time. On the construction side, existing methods range from iterative refinement through execution feedback~\citep{wang2026skillx} to reinforcement-learning-based skill discovery~\citep{kang2025sage, xia2026skillrl} and parameter-level skill acquisition via LoRA merging~\citep{ye2024cycleqd, liu2025psec}. On the routing side, SkillRouter~\citep{zheng2026skillrouter} shows that metadata-only retrieval is insufficient for large skill libraries and proposes a retrieval-reranking pipeline, while others study capacity limits of single-agent routing~\citep{li2026singleagent} and the difficulty of skill retrieval from uncurated libraries~\citep{zhang2026benchmarking}. These efforts expand the supply of skills and improve their acquisition and selection. Instruction-level composition additionally requires analyzing interactions among independently authored directives and reconciling them into a shared operational specification.

\paragraph{Skill composition.}
The most related work attempts to compose multiple skills together.~\citet{jiao2026agenticproposing} dynamically select and compose reasoning skills during problem synthesis, while \citet{meng2026constraintcomposition} define structured skill libraries with precondition and post-effect checking. At the parameter level,~\citet{prabhakar2025lora} and~\citet{zhang2025mergevla} merge LoRA modules across skills. However, these methods operate either at the planning level (selecting which skill to invoke next) or at the weight level (merging adapter parameters directly). None of them address instruction-level reconciliation: given two \texttt{SKILL.md} files with partially overlapping and potentially conflicting natural-language directives, how should they be merged into one coherent and executable specification?

\paragraph{Multi-agent verification.}
Recent work shows that multi-agent interaction improves single-model reliability. LLM-as-judge frameworks~\citep{zheng2023judging} use one model to evaluate another's output, and multi-agent debate~\citep{liang2024encouraging} demonstrates that adversarial critique reduces hallucination. \Name applies this principle: a Classifier Agent and an Adversarial Verifier Agent form a debate protocol for conflict classification, re-examining uncertain judgments against the original pair of source constraints (Section~\ref{sec:detection}; threshold analysis in Appendix~\ref{app:threshold_sweep}).

\paragraph{Instruction-following under constraint pressure.}
Independent evidence from the instruction-following literature confirms that the problem is real. ComplexBench~\citep{wen2024complexbench} shows that multi-constraint composition is a distinct challenge for frontier LLMs. RECAST~\citep{guo2025recast} demonstrates sharp performance degradation as the number of explicit requirements grows. ConInstruct~\citep{qian2025coninstruct} adds a particularly relevant finding: even strong models often fail to surface or clarify conflicts, instead responding under partial or silently compromised compliance. These results establish that implicit runtime reconciliation is unreliable under constraint-dense conditions, which is precisely the regime that unresolved multi-skill loading creates. \Name addresses this gap through explicit, offline reconciliation of source directives and their activation conditions, producing a reusable instruction set that is shared across tasks and executor models.\looseness=-1

\suppressfloats[t]

\section{Method}
\label{sec:method}

\subsection{Problem Setup}
\label{sec:formalization}

\begin{definition}[Agent Skill]
\label{def:skill}
An \emph{agent skill} is a tuple $s=(m,I,P,R)$, where $m=(\text{name},\text{desc})$ is metadata for discovery, $I=\{d_1,\dots,d_n\}$ is a set of natural-language directives, $P$ is a set of executable scripts, and $R$ is a set of auxiliary resources such as templates and documentation.
\end{definition}

\begin{figure*}[t]
\centering

\includegraphics[width=\textwidth]{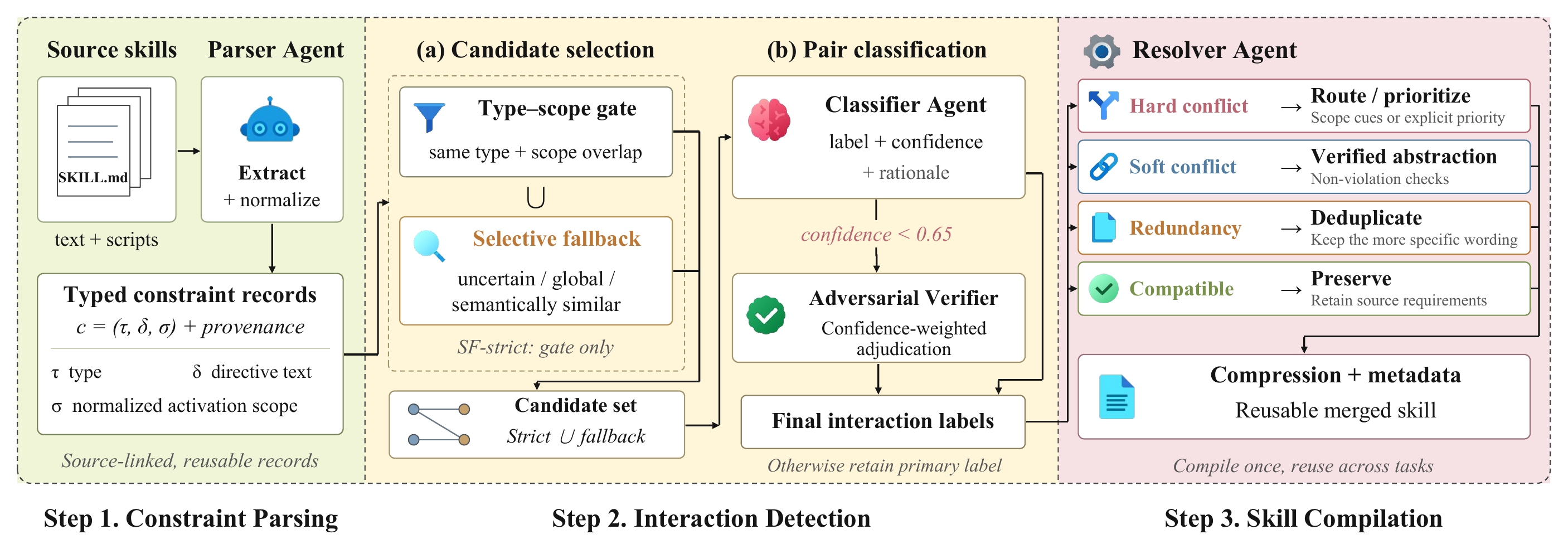}

\caption{
Overview of \Name.
A Parser Agent extracts typed constraints from each skill. Candidate generation combines a type--scope gate with selective fallback for potentially missed interactions. A Classifier Agent and Adversarial Verifier Agent classify surviving pairs through a debate protocol. A Resolver Agent applies type-dispatched compilation (routing, abstraction, or deduplication) and compresses the result into a unified merged skill.
}
\label{fig:skillfusion_pipeline}

\end{figure*}

Metadata enables discovery; instructions and scripts encode the commitments to be composed.

\begin{samepage}
\begin{definition}[Typed Constraint]
\label{def:constraint}
A typed constraint $c=(\tau,\delta,\sigma)$ consists of a behavioral type $\tau$, directive text $\delta$, and activation scope $\sigma$, with
\[
\tau \in \{\textsc{format},\textsc{behavior},\textsc{tool},\textsc{priority},\textsc{style}\}.
\]
\end{definition}
\end{samepage}

Intuitively, the type $\tau$ indicates \emph{what behavioral dimension} a directive governs (e.g., output shape, procedural step, or tool-use rule), while the scope $\sigma$ indicates precisely \emph{when} it applies. A global scope~($\sigma=\top$) means the directive is unconditional. Otherwise, it is active only when the current task context matches its scope cues. For a skill $s$, we write $\mathcal{C}(s)=\{c_1,\dots,c_m\}$ to denote the full set of typed constraints extracted from $I$ and $P$.

Given skills $s_A,s_B$, the compiler produces a merged skill $s_{AB}$ that retains compatible requirements and specifies how conflicting directives are resolved. Naive concatenation, $I_A\oplus I_B$, provides the unresolved reference. We assess the merged skill through task Success, instruction Compliance, and source-constraint Preservation.

\paragraph{Interaction types.}
We distinguish hard conflicts, soft conflicts, and redundancy as introduced in Section~\ref{sec:intro}. Hard conflicts require incompatible actions under overlapping scopes; soft conflicts compete without logical inconsistency; redundancy repeats an existing requirement. Section~\ref{sec:mechanistic} evaluates their isolated and joint effects in controlled instruction combinations.

\subsection{\Name Pipeline}
\label{sec:overview}

\Name uses four LLM roles across three stages (Figure~\ref{fig:skillfusion_pipeline}): parsing, candidate interaction detection with adversarial re-judging, and resolution with compression. It outputs a merged \texttt{SKILL.md}, discovery metadata, and preserved scripts and resources. Detailed prompts and implementation procedures appear in Appendices~\ref{app:prompt_templates} and~\ref{app:pipeline_details}.

\subsubsection{Stage~I: Constraint Parsing}
\label{sec:parsing}

Skill files vary widely in style: declarative checklists, narrative prose, and helper scripts all encode behavioral assumptions. Each skill must first be mapped into the representation of Definition~\ref{def:constraint}. An LLM-based Parser Agent $\mathcal{P}$ decomposes each instruction file $I$ into minimal directives:
\[
\mathcal{P}(I)=\{(\tau_1,\delta_1,\sigma_1),\dots,(\tau_m,\delta_m,\sigma_m)\}.
\]
The parser uses a structured JSON schema enforcing minimality and scope accuracy.
Because raw scope language is often informal (e.g., ``for spreadsheet tasks''), we normalize each scope into structured cues (task category, workflow phase, file pattern, tool context) while retaining original wording for reconstruction. Behaviorally important constraints may also be encoded implicitly in scripts; for example, a workbook-editing script may enforce preservation of existing cell formulas. We extract such constraints via static inspection followed by a two-pass semantic filter. At the end of parsing, each skill is a finite set of typed, scoped constraints with provenance links.

\subsubsection{Stage~II: Conflict Detection}
\label{sec:detection}

The goal of Stage~II is to determine how the constraints of two parsed skills interact. Comparing all cross-skill pairs directly would be both expensive and noisy, so we adopt a two-phase procedure: graph-based pruning followed by an adversarial debate protocol.

\paragraph{Constraint Interaction Graph.}
For each pair of source skills, we form a bipartite constraint graph; larger pools include all skill pairs. The strict edge set uses the type--scope gate:
\[
E_{\mathrm{strict}}=\{(c_i,c_j):\tau_i=\tau_j\ \wedge\ \sigma_i\cap\sigma_j\neq\varnothing\}.
\]
Typing and scope estimates can omit genuine interactions. The final system adds a selective fallback set $E_{\mathrm{fb}}$: excluded pairs are reconsidered when typing or scope is uncertain, either scope is global, or semantic similarity is high. Classification operates on $E=E_{\mathrm{strict}}\cup E_{\mathrm{fb}}$. We call this configuration \textbf{SF-final}; \textbf{SF-strict} uses $E_{\mathrm{fb}}=\varnothing$. SF-final admits 7.8--19.0\% of possible cross-skill pairs in the three benchmark pools (Appendix~\ref{app:pipeline_analysis}).

\paragraph{Adversarial debate classification.}
For each surviving edge, a \emph{Classifier Agent} $\mathcal{J}_1$ labels the interaction into one of four categories:
\[
\oplus \;(\text{hard conflict}), \quad
\ominus \;(\text{soft conflict}), \quad
\equiv \;(\text{redundancy}), \quad
\perp \;(\text{compatibility}).
\]
The classifier receives both constraints as structured JSON records and returns a label, confidence score, and rationale.
To guard against over-classification, an \emph{Adversarial Verifier Agent} $\mathcal{J}_2$ re-evaluates low-confidence verdicts ($\gamma<0.65$) under a skeptical counter-prompt.
A confidence-weighted rule adjudicates disagreements (Algorithm~\ref{alg:detection}).

\begin{algorithm}[t]
\caption{Candidate Generation and Adversarial Conflict Detection}
\label{alg:detection}
\footnotesize
\textbf{Input:} $\mathcal{C}(s_A)$, $\mathcal{C}(s_B)$; threshold $\theta$ \quad \textbf{Output:} Labeled set $\mathcal{L}$ \\[1pt]
\textit{// Phase 1: Construct Constraint Interaction Graph} \\
1: $E \leftarrow E_{\mathrm{strict}}\cup E_{\mathrm{fb}}$ \hfill $\triangleright$ \textit{Selective fallback enabled in SF-final} \\
\textit{// Phase 2: Adversarial debate classification} \\
2: $\mathcal{L} \leftarrow \varnothing$ \\
3: \textbf{for all} $(c_i, c_j) \in E$ \textbf{do} \\
4: \quad $(l_1, \gamma_1) \leftarrow \mathcal{J}_1(c_i, c_j)$ \hfill $\triangleright$ \textit{Classifier Agent} \\
5: \quad \textbf{if} $\gamma_1 \geq \theta$: \; $\mathcal{L} \leftarrow \mathcal{L} \cup \{(c_i, c_j, l_1)\}$ \\
6: \quad \textbf{else}: $(l_2, \gamma_2) \leftarrow \mathcal{J}_2(c_i, c_j)$ \hfill $\triangleright$ \textit{Adversarial Verifier} \\
7: \quad\quad $l^{*} \leftarrow \textsc{Adjudicate}(l_1, \gamma_1, l_2, \gamma_2)$; \; $\mathcal{L} \leftarrow \mathcal{L} \cup \{(c_i, c_j, l^{*})\}$ \\
8: \textbf{return} $\mathcal{L}$
\end{algorithm}

\subsubsection{Stage~III: Conflict Resolution and Compression}
\label{sec:resolution}
In Stage~III, the \emph{Resolver Agent} compiles detected interactions
into a merged instruction set. This compilation is governed by the
goal of \emph{behavioral preservation}: retained and synthesized directives maintain links to their sources. Compatible requirements are preserved or subsumed; genuinely incompatible requirements require an explicit routing or priority decision.

For task contexts within separable source scopes, the resolver defines:
\[
\delta_{ij}^{\mathrm{merge}}(t)=
\begin{cases}
\delta_i, & \text{if } \phi_{ij}(t),\\
\delta_j, & \text{otherwise},
\end{cases}
\]
where $t$ is the task context and $\phi_{ij}$ derives from normalized scope cues. If no reliable discriminator exists, \Name falls back to an explicit disambiguation block that specifies a priority and fallback. For soft conflicts (competing but not contradictory), the resolver derives a super-directive $\delta_{ij}^{\mathrm{abs}}$ that satisfies \emph{subsumption} and \emph{specificity}. Each candidate is verified by pairwise non-violation checks against the original constraints, with a fallback: if either check fails, the hard-conflict resolver chooses routing or explicit priority. For redundant pairs, the resolver keeps the more specific formulation
and drops the weaker duplicate.

After resolution, three compression operators are applied: shared-directive
factoring, entailment-verified thematic clustering, and structural
\looseness=-1
normalization. These operators reduce repetition while checking synthesized directives against their sources. Finally, the compiler synthesizes metadata $m_{AB}$ covering both source skills. Activation recall measures whether the merged description preserves discovery coverage.

\section{Experiments}
\label{sec:experiments}
\subsection{Experimental Settings}
\label{sec:settings}
\paragraph{Benchmarks and skill pools.}
We evaluate on SpreadsheetBench Verified~\citep{ma2024spreadsheetbench} (400 tasks, hard restriction), WebArena-Verified~\citep{zhou2024webarena} (all 812 tasks), and SWE-bench Verified~\citep{jimenez2023swe,openai2024swebenchverified} (500 tasks). Each is paired with five independently authored public skills covering spreadsheet workflows, browser interaction, or software engineering. HumanEval+~\citep{liu2023evalplus} (164 tasks) supplies the controlled testbed, with synthetic skills constructed from 12 coding-practice directives. Appendix~\ref{app:skill_pools} lists skill sources and construction details.

\paragraph{Baselines and metrics.}
We compare No Skill, retrospective Single-Skill Oracle, Multi-Skill Loading (MSL), Naive Concat, one-call LLM Rewrite, Skill Routing~\citep{zheng2026skillrouter}, and three-call Rewrite--Critique--Revise (RCR). The oracle selects the best single-skill outcome per task and is a reference within that candidate family. SF-strict uses only the type--scope gate; SF-final, our full method, adds selective fallback with verifier threshold .65.
Success is benchmark-native correctness (pass@1 on HumanEval+). Compliance is the fraction of supplied instructions judged satisfied~\citep{zheng2023judging}; instruction sets vary by method, so we report it alongside Success and Preservation. Compression is $1-\text{output tokens}/\text{input tokens}$ over instruction bodies. Ablations measure synthetic conflict recall, source Preservation, and metadata Activation recall.

\paragraph{Protocol.}
The compiler uses DeepSeek-V3.2~\citep{deepseek2025v3} at temperature 0.0 through CAMEL~\citep{li2023camel}. One fixed artifact per method and skill pool is reused across tasks and transferred to GLM-4.7~\citep{glm2025glm4} and MiniMax-M2.5~\citep{minimax2026m25}. Every main and cross-model column uses three full evaluations. HumanEval+ generates one program per task per run: four runs for ablations and gate/threshold tests, and 20 for scaling and mechanism scenarios. Intervals show Student-$t$ 95\% confidence intervals for the mean across repeated executions on a fixed task set. Difference intervals use within-block paired score differences. Appendix~\ref{app:new_experiments} details record reuse, denominators, cost, and full results.

\subsection{Main Results}
\label{sec:main_results}
\begin{figure}[!htbp]
\centering\includegraphics[width=\linewidth]{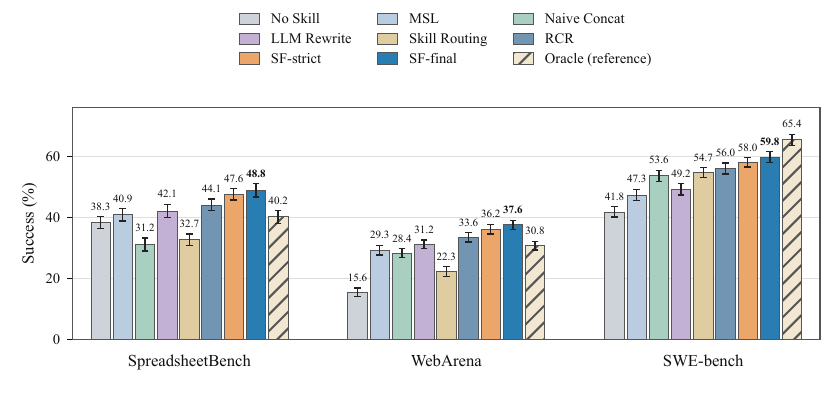}
\caption{Detailed main results across three fixed task sets. Bars and labels show mean Success (\%); error bars show Student-$t$ 95\% CIs over three complete runs. All benchmarks share the same scale. Hatched bars denote the retrospective oracle reference.}
\label{fig:main_results}
\end{figure}

By explicitly reconciling interactions, SF-final achieves the highest non-oracle Success on all three benchmarks (Figure~\ref{fig:main_results}), improving over three-call RCR by 4.8, 4.0, and 3.8 percentage points. All three paired 95\% intervals exclude zero (Appendix~\ref{app:repeated_results}). The advantage persists over a baseline that already critiques and revises its merge; the ablations identify the contributing composition decisions.
\begin{table}[tbp]
\centering
\caption{Main-benchmark Compliance (C) and compression (R), in percent. Results cover all nine methods on the three benchmarks. Compliance is measured against each method's supplied instruction set. Dashes indicate inapplicable metrics.}
\label{tab:main_diagnostics}
\small\setlength{\tabcolsep}{8pt}\renewcommand{\arraystretch}{1.10}
\begin{tabular}{@{}lrrrrrr@{}}
\toprule
& \multicolumn{2}{c}{SpreadsheetBench} & \multicolumn{2}{c}{WebArena} & \multicolumn{2}{c}{SWE-bench} \\
\cmidrule(lr){2-3}\cmidrule(lr){4-5}\cmidrule(lr){6-7}
Method & C & R & C & R & C & R \\
\midrule
No Skill & -- & -- & -- & -- & -- & -- \\
Single-Skill Oracle & 53.4 & -- & 77.1 & -- & 99.1 & -- \\
MSL & 48.7 & 0.0 & 74.9 & 0.0 & 99.3 & 0.0 \\
Naive Concat & 62.1 & 0.0 & 78.3 & 0.0 & 100.0 & 0.0 \\
LLM Rewrite & 57.3 & 22.3 & 79.4 & 38.9 & 99.7 & 15.8 \\
Skill Routing & 46.2 & -- & 76.2 & -- & 99.6 & -- \\
RCR & 59.6 & 25.5 & 80.6 & 34.2 & 99.3 & 14.1 \\
SF-strict & 60.8 & 37.1 & 81.7 & 17.4 & 99.4 & 10.3 \\
\midrule\rowcolor{tab-highlight} SF-final & 62.3 & 36.0 & 83.1 & 18.0 & 99.5 & 10.5 \\
\bottomrule
\end{tabular}
\end{table}

The baselines expose the composition difficulty. On SpreadsheetBench, Naive Concat falls below No Skill, whereas SF-final exceeds even the retrospective single-skill oracle. The same oracle comparison holds on WebArena, supporting the value of combining complementary skills through reconciliation. SWE-bench retains a stronger oracle, indicating additional value in task-specific selection. SF-final also compresses less than RCR on WebArena and SWE-bench (Table~\ref{tab:main_diagnostics}). Its higher Success favors selective preservation over maximizing text reduction. Compilation cost is incurred once per skill pool and amortized through artifact reuse (Appendix~\ref{app:repeated_results}).
\subsection{Component Ablations}
\label{sec:component_ablation}
\begin{figure}[!htbp]
\centering\includegraphics[width=0.9\linewidth]{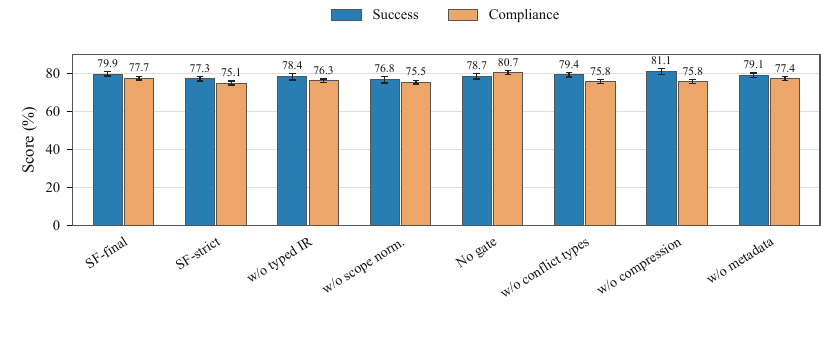}
\caption{HumanEval+ S60 component ablations. Bars show means and error bars show Student-$t$ 95\% CIs over four runs on 164 fixed tasks. Additional recall, preservation, activation, and compression diagnostics appear in Appendix~\ref{app:ablation_full}.}
\label{fig:ablation}
\end{figure}

The S60 ablations separate representation, resolution, and packaging (Figure~\ref{fig:ablation}; Appendix~\ref{app:ablation_full}). Removing scope normalization gives the largest Success drop, from .799 to .768, and also lowers recall and Preservation. Typed representation shows the same direction. Scope thus matters at both candidate selection and source-requirement retention.

Removing conflict-type distinction leaves recall at .833 but reduces Preservation from .900 to .850, locating its contribution in resolution after detection. Removing metadata synthesis leaves Preservation unchanged but lowers Activation recall, separating discovery from instruction retention. Compression improves Compliance and reduces length, while its removal yields higher mean Success (.811): compactness carries a task-performance trade-off in this setting.

\Needspace{6\baselineskip}
\subsection{Human Audit and Candidate Robustness}
\label{sec:robustness}
The human audit identifies early pruning as a recoverable bottleneck. SF-strict misses 24 of 60 incompatible pairs; selective fallback recovers 14, largely from typing, scope normalization, and cross-type interactions. F1 rises from .713 to .840, while precision decreases from .878 to .847. Revisiting uncertain exclusions therefore recovers useful interactions that a strict structural gate would permanently hide from the resolver (Appendix~\ref{app:gate_audit}).
\begin{table}[tbp]
\centering
\caption{Candidate-generation trade-offs. P/R/F1 use the 300-pair human audit; candidates and API cost use a separate 5,216-pair calibration set. Preservation and Success use HumanEval+ S60; Spreadsheet Success uses the main benchmark. Bold marks best scores and lowest API cost.}
\label{tab:gate_main}
\footnotesize
\setlength{\tabcolsep}{4pt}
\renewcommand{\arraystretch}{1.12}
\begin{tabular}{@{}lcccccc@{}}
\toprule
Variant & P / R / F1 & Candidates & API & Pres. & S60 & Sheet \\
\midrule
SF-strict & \textbf{0.878}/0.600/0.713 & 612 & \textbf{1.000} & 0.883 & 0.773 & 0.476 \\
No gate & 0.696/\textbf{0.917}/0.791 & 5,216 & 7.610 & 0.717 & 0.787 & 0.457 \\
\rowcolor{tab-highlight} SF-final & 0.847/0.833/\textbf{0.840} & 1,184 & 1.820 & \textbf{0.900} & \textbf{0.799} & \textbf{0.488} \\
\bottomrule
\end{tabular}
\end{table}

Exhaustive comparison exposes the complementary failure mode (Table~\ref{tab:gate_main}). No gate has the highest recall, but generates 24 false positives versus nine for SF-final and lowers Preservation and task Success. This is consistent with unnecessary conflict resolutions altering otherwise useful directives. SF-final also uses a much lower calibration API ratio (1.820 versus 7.610). Together, these results favor selective recovery: retain structural filtering while reopening plausible misses. The verifier sweep changes Success by at most .006 across the two testbeds, suggesting that the observed performance is relatively stable over the tested threshold range (Appendix~\ref{app:threshold_sweep}).
\subsection{Cross-Model Generalization}
\label{sec:cross_model}
SF-final leads all non-oracle methods in the six benchmark--backbone combinations, with gains over RCR of 1.5--2.9 points on the two transfer backbones (Appendix Figures~\ref{fig:cross_spreadsheetbench} and~\ref{fig:cross_webarena}). These fixed artifacts transfer without backbone-specific recompilation. Free-form rewriting is less consistent: its WebArena Success falls to .098 with MiniMax, while SF-final retains .236. Explicitly resolved instructions thus support reuse across the tested executors despite differing absolute performance.
\subsection{Scaling with Instruction Complexity}
\label{sec:scaling}
\begin{figure}[!htbp]
\centering\includegraphics[width=0.9\linewidth]{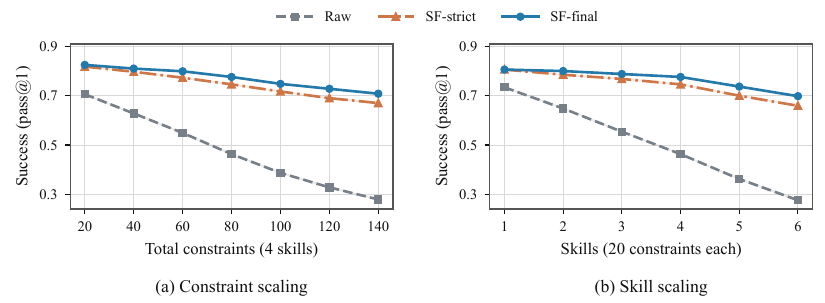}
\caption{HumanEval+ Success as instruction complexity increases. Left: four skills with varying constraint counts. Right: 20 constraints per skill. Each point uses 164 tasks across 20 runs; shading shows Student-$t$ 95\% confidence intervals for mean scores on the fixed task set. The 80-constraint and four-skill points reuse the same records.}\label{fig:scaling}
\end{figure}

The benefit of compilation grows with instruction complexity (Figure~\ref{fig:scaling}). Raw loads the unresolved instruction combination. With four skills, increasing constraints from 20 to 140 reduces Raw Success from .706 to .279, while SF-final declines from .825 to .708. Varying the number of skills shows the same pattern; final--Raw gains reach 42.9 and 42.2 points at the two endpoints. Thus, reconciliation becomes more valuable as the executor faces a larger unresolved instruction set. The final--strict gap also widens along both axes, linking selective recovery to robustness under increasing composition demands. Appendix~\ref{app:scaling_full} reports Compliance and paired gain intervals.

\subsection{Controlled Mechanism Analysis}
\label{sec:mechanistic}
Unresolved skill interactions reduce execution quality (Table~\ref{tab:mechanistic}). The hard-conflict condition has the largest isolated Success loss; redundant instructions and competing preferences also lower Success and Compliance. Redundancy even exceeds the soft-conflict loss in these scenarios. These distinct burdens motivate routing, abstraction, and deduplication, whose differentiated treatment is supported by the conflict-type ablation.

The joint condition yields the lowest Success and Compliance. Its mean Success drop is .027 smaller than the sum of the isolated drops (.219), so isolated penalties do not simply add when interaction types coexist. Joint evaluation therefore complements the isolated comparisons.
\begin{table}[!htbp]
\centering
\caption{Unresolved interaction conditions on HumanEval+. Scores are mean Success and Compliance over 20 runs on 164 fixed tasks. The joint condition combines all three interaction types. Bold marks the highest mean in each column.}
\label{tab:mechanistic}
\small\setlength{\tabcolsep}{12pt}\renewcommand{\arraystretch}{1.08}
\begin{tabular}{@{}lrr@{}}
\toprule
Condition & Success & Compliance \\
\midrule
Single-skill reference & \textbf{0.823} & \textbf{0.887} \\
\midrule
Hard conflict & 0.716 & 0.745 \\
Redundancy / dilution & 0.758 & 0.802 \\
Soft conflict / ambiguity & 0.776 & 0.816 \\
Joint condition & 0.631 & 0.654 \\
\bottomrule
\end{tabular}
\end{table}

\section{Conclusion}
\looseness=-1
\Name compiles independently authored skills into reusable specifications through typed, scoped interaction analysis. Interaction-specific resolution improves task Success across three agent benchmarks, and selective fallback recovers conflicts missed by strict pruning. Human auditing, component ablations, and controlled mechanism experiments connect these gains to concrete stages of the compiler. Cross-model transfer and instruction scaling further demonstrate the utility of resolving skill interactions before execution.

\section*{Ethics Statement}
This work studies skill composition using public agent benchmarks and reusable skill instructions. More effective composition can also facilitate harmful or unauthorized workflows, and LLM-generated resolutions can introduce incorrect instructions. Deployment should therefore retain independent permission checks, sandboxed execution, and human approval for consequential actions. Reuse and redistribution should respect benchmark terms and upstream skill licenses. Private files, credentials, and personal data require appropriate authorization and protection when third-party model APIs are used. Our evaluation concerns task performance and instruction consistency; operational safety requires separate assessment.

\section*{AI Use Statement}
We used GPT-6 Astra to assist with grammar polishing and revision.

\section*{Reproducibility Statement}
The appendices provide evaluation protocols, supplementary results, compilation costs, human-audit counts, prompt templates, and skill-pool details. The code will be open-sourced after the review process concludes.

\bibliography{references}

\begin{thebibliography}{44}
\providecommand{\natexlab}[1]{#1}
\providecommand{\url}[1]{\texttt{#1}}
\expandafter\ifx\csname urlstyle\endcsname\relax
  \providecommand{\doi}[1]{doi: #1}\else
  \providecommand{\doi}{doi: \begingroup \urlstyle{rm}\Url}\fi

\bibitem[{Anthropic}(2026)]{anthropic2026skillsguide}
{Anthropic}.
\newblock The complete guide to building skills for claude.
\newblock
  \url{https://resources.anthropic.com/hubfs/The-Complete-Guide-to-Building-Skill-for-
  Claude.pdf}, 2026.

\bibitem[{DeepSeek-AI} et~al.(2025){DeepSeek-AI}, Liu, Mei,
  et~al.]{deepseek2025v3}
{DeepSeek-AI}, Aixin Liu, Aoxue Mei, et~al.
\newblock {DeepSeek}-{V3.2}: Pushing the frontier of open large language
  models.
\newblock \emph{arXiv preprint arXiv:2512.02556}, 2025.

\bibitem[Fu et~al.(2026)Fu, Zhang, Zhang, Wang, Huang, and
  Luo]{zhang2025mergevla}
Yuxia Fu, Zhizhen Zhang, Yuqi Zhang, Zijian Wang, Zi~Huang, and Yadan Luo.
\newblock {MergeVLA}: Cross-skill model merging toward a generalist
  vision-language-action agent.
\newblock \emph{arXiv preprint arXiv:2511.18810}, 2026.

\bibitem[Guo et~al.(2025)Guo, Liu, Xie, Xu, Huang, Tian, Xu, Shen,
  et~al.]{guo2025recast}
Zhengkang Guo, Wenhao Liu, Mingchen Xie, Jingwen Xu, Zisu Huang, Muzhao Tian,
  Jianhan Xu, Yuanzhe Shen, et~al.
\newblock {RECAST}: Expanding the boundaries of {LLM}s' complex instruction
  following with multi-constraint data.
\newblock \emph{arXiv preprint arXiv:2505.19030}, 2025.

\bibitem[He et~al.(2026)He, Zhang, Chen, Chen, Yu, Yuan, and
  Yiu]{qian2025coninstruct}
Xingwei He, Qianru Zhang, Pengfei Chen, Guanhua Chen, Linlin Yu, Yuan Yuan, and
  Siu-Ming Yiu.
\newblock Coninstruct: Evaluating large language models on conflict detection
  and resolution in instructions.
\newblock In \emph{AAAI}, 2026.

\bibitem[Jiao et~al.(2026)Jiao, Wang, Zhang, Ren, Wang, Zhao, Wei, and
  Zhang]{jiao2026agenticproposing}
Zhengbo Jiao, Shaobo Wang, Zifan Zhang, Xuan Ren, Wei Wang, Bing Zhao, Hu~Wei,
  and Linfeng Zhang.
\newblock Agentic proposing: Enhancing large language model reasoning via
  compositional skill synthesis.
\newblock \emph{arXiv preprint arXiv:2602.03279}, 2026.

\bibitem[Jimenez et~al.(2023)Jimenez, Yang, Wettig, Yao, Pei, Press, and
  Narasimhan]{jimenez2023swe}
Carlos~E. Jimenez, John Yang, Alexander Wettig, Shunyu Yao, Kexin Pei, Ofir
  Press, and Karthik Narasimhan.
\newblock Swe-bench: Can language models resolve real-world github issues?
\newblock In \emph{The twelfth international conference on learning
  representations}, 2023.

\bibitem[Kuroki et~al.(2025)Kuroki, Nakamura, Akiba, and Tang]{ye2024cycleqd}
So~Kuroki, Taishi Nakamura, Takuya Akiba, and Yujin Tang.
\newblock Agent skill acquisition for large language models via {CycleQD}.
\newblock In \emph{ICLR}, 2025.

\bibitem[Li et~al.(2023)Li, Hammoud, Itani, Khizbullin, and
  Ghanem]{li2023camel}
Guohao Li, Hasan Abed Al~Kader Hammoud, Hani Itani, Dmitrii Khizbullin, and
  Bernard Ghanem.
\newblock {CAMEL}: Communicative agents for ``mind'' exploration of large
  language model society.
\newblock In \emph{NeurIPS}, 2023.

\bibitem[Li(2026)]{li2026singleagent}
Xiaoxiao Li.
\newblock When single-agent with skills replace multi-agent systems and when
  they fail.
\newblock \emph{arXiv preprint arXiv:2601.04748}, 2026.

\bibitem[Liang et~al.(2024)Liang, He, Jiao, Wang, Wang, Wang,
  et~al.]{liang2024encouraging}
Tian Liang, Zhiwei He, Wenxiang Jiao, Xing Wang, Yan Wang, Rui Wang, et~al.
\newblock Encouraging divergent thinking in large language models through
  multi-agent debate.
\newblock In \emph{Proceedings of the 2024 conference on empirical methods in
  natural language processing}, 2024.

\bibitem[Ling et~al.(2026)Ling, Zhong, and Huang]{ling2026agentskills}
George Ling, Shanshan Zhong, and Richard Huang.
\newblock Agent skills: A data-driven analysis of claude skills for extending
  large language model functionality.
\newblock \emph{arXiv preprint arXiv:2602.08004}, 2026.

\bibitem[Liu et~al.(2023)Liu, Xia, Wang, and Zhang]{liu2023evalplus}
Jiawei Liu, Chunqiu~Steven Xia, Yuyao Wang, and Lingming Zhang.
\newblock Is your code generated by {ChatGPT} really correct? rigorous
  evaluation of large language models for code generation.
\newblock In \emph{Advances in Neural Information Processing Systems
  (NeurIPS)}, 2023.

\bibitem[Liu et~al.(2025)Liu, Li, Zheng, Niu, Lan, Xu, and Zhan]{liu2025psec}
Tenglong Liu, Jianxiong Li, Yinan Zheng, Haoyi Niu, Yixing Lan, Xin Xu, and
  Xianyuan Zhan.
\newblock Skill expansion and composition in parameter space.
\newblock In \emph{ICLR}, 2025.

\bibitem[Liu et~al.(2024)Liu, Yu, Zhang, Xu, Lei, Lai, Gu, Ding, Men, Yang,
  et~al.]{liu2024agentbench}
Xiao Liu, Hao Yu, Hanchen Zhang, Yifan Xu, Xuanyu Lei, Hanyu Lai, Yu~Gu,
  Hangliang Ding, Kaiwen Men, Kejuan Yang, et~al.
\newblock {AgentBench}: Evaluating {LLM}s as agents.
\newblock In \emph{ICLR}, 2024.

\bibitem[Liu et~al.(2026)Liu, Ji, An, Jaakkola, Zhang, and
  Chang]{zhang2026benchmarking}
Yujian Liu, Jiabao Ji, Li~An, Tommi Jaakkola, Yang Zhang, and Shiyu Chang.
\newblock How well do agentic skills work in the wild: Benchmarking {LLM} skill
  usage in realistic settings.
\newblock \emph{arXiv preprint arXiv:2604.04323}, 2026.

\bibitem[Ma et~al.(2024)Ma, Zhang, Zhang, Yu, Zhang, Zhang, Luo, Wang, and
  Tang]{ma2024spreadsheetbench}
Zeyao Ma, Bohan Zhang, Jing Zhang, Jifan Yu, Xiaokang Zhang, Xiaohan Zhang,
  Sijia Luo, Xi~Wang, and Jie Tang.
\newblock {SpreadsheetBench}: Towards challenging real world spreadsheet
  manipulation.
\newblock In \emph{Advances in Neural Information Processing Systems (NeurIPS),
  Datasets and Benchmarks Track}, 2024.
\newblock URL
  \url{http://papers.nips.cc/paper_files/paper/2024/hash/ac840df270ac537dd74530a15c332684-Abstract-Datasets_and_Benchmarks_Track.html}.

\bibitem[Meng et~al.(2026)Meng, Huang, Zhang, Yang, Zeng, and
  Huang]{meng2026constraintcomposition}
Ru~Meng, Shaoyu Huang, Xinyue Zhang, Yourong Yang, Kewei Zeng, and Zhibin
  Huang.
\newblock Constraint-consistent skill composition for reliable zero-shot task
  generalization in llm agents.
\newblock \emph{2026 9th International Symposium on Big Data and Applied
  Statistics (ISBDAS)}, 2026.

\bibitem[{MiniMax}(2026)]{minimax2026m25}
{MiniMax}.
\newblock {MiniMax M2.5}: Built for real-world productivity.
\newblock \url{https://www.minimax.io/news/minimax-m25}, 2026.

\bibitem[{OpenAI}(2024)]{openai2024swebenchverified}
{OpenAI}.
\newblock Introducing {SWE-bench} verified.
\newblock \url{https://openai.com/index/introducing-swe-bench-verified/}, 2024.
\newblock Human-validated 500-task subset of \citet{jimenez2023swe}.

\bibitem[{OpenAI}(2026)]{openai2026codex}
{OpenAI}.
\newblock Codex skills.
\newblock \url{https://developers.openai.com/codex/skills/}, 2026.

\bibitem[Patil et~al.(2024)Patil, Zhang, Wang, and Gonzalez]{patil2023gorilla}
Shishir~G. Patil, Tianjun Zhang, Xin Wang, and Joseph~E. Gonzalez.
\newblock Gorilla: Large language model connected with massive {API}s.
\newblock In \emph{NeurIPS}, 2024.

\bibitem[Prabhakar et~al.(2025)Prabhakar, Li, Narasimhan, Kakade, Malach, and
  Jelassi]{prabhakar2025lora}
Akshara Prabhakar, Yuanzhi Li, Karthik Narasimhan, Sham Kakade, Eran Malach,
  and Samy Jelassi.
\newblock {LoRA} soups: Merging {LoRA}s for practical skill composition tasks.
\newblock In \emph{COLING}, 2025.

\bibitem[Qin et~al.(2024)Qin, Liang, Ye, Zhu, Yan, Lu, Lin, Cong,
  et~al.]{qin2024toolbench}
Yujia Qin, Shihao Liang, Yining Ye, Kunlun Zhu, Lan Yan, Yaxi Lu, Yankai Lin,
  Xin Cong, et~al.
\newblock {ToolLLM}: Facilitating large language models to master 16000+
  real-world {API}s.
\newblock In \emph{ICLR}, 2024.

\bibitem[Schick et~al.(2023)Schick, Dwivedi-Yu, Dess{\`i}, Raileanu, Lomeli,
  et~al.]{schick2023toolformer}
Timo Schick, Jane Dwivedi-Yu, Roberto Dess{\`i}, Roberta Raileanu, Maria
  Lomeli, et~al.
\newblock Toolformer: Language models can teach themselves to use tools.
\newblock In \emph{NeurIPS}, 2023.

\bibitem[Sera(2026)]{openai2026skillsblog}
Kazuhiro Sera.
\newblock Using skills to accelerate oss maintenance.
\newblock \url{https://developers.openai.com/blog/skills-agents-sdk}, 2026.

\bibitem[Shen et~al.(2023)Shen, Song, Tan, Li, Lu, and
  Zhuang]{shen2023hugginggpt}
Yongliang Shen, Kaitao Song, Xu~Tan, Dongsheng Li, Weiming Lu, and Yueting
  Zhuang.
\newblock {HuggingGPT}: Solving {AI} tasks with {ChatGPT} and its friends in
  {Hugging Face}.
\newblock In \emph{NeurIPS}, 2023.

\bibitem[Shinn et~al.(2023)Shinn, Cassano, Gopinath, Narasimhan, and
  Yao]{shinn2023reflexion}
Noah Shinn, Federico Cassano, Ashwin Gopinath, Karthik Narasimhan, and Shunyu
  Yao.
\newblock Reflexion: Language agents with verbal reinforcement learning.
\newblock In \emph{NeurIPS}, 2023.

\bibitem[Tran et~al.(2025)Tran, Dao, Nguyen, Pham, O'Sullivan, and
  Nguyen]{guo2025multiagent}
Khanh-Tung Tran, Dung Dao, Minh-Duong Nguyen, Quoc-Viet Pham, Barry O'Sullivan,
  and Hoang~D. Nguyen.
\newblock Multi-agent collaboration mechanisms: A survey of {LLM}s.
\newblock \emph{arXiv preprint arXiv:2501.06322}, 2025.

\bibitem[Wang et~al.(2026)Wang, Yu, Xie, Yao, Fang, Qiao, Cao, Zheng,
  et~al.]{wang2026skillx}
Chenxi Wang, Zhuoyun Yu, Xin Xie, Wuguannan Yao, Runnan Fang, Shuofei Qiao,
  Kexin Cao, Guozhou Zheng, et~al.
\newblock {SkillX}: Automatically constructing skill knowledge bases for
  agents.
\newblock \emph{arXiv preprint arXiv:2604.04804}, 2026.

\bibitem[Wang et~al.(2023)Wang, Xie, Jiang, Mandlekar, Xiao, Zhu, Fan, and
  Anandkumar]{wang2023voyager}
Guanzhi Wang, Yuqi Xie, Yunfan Jiang, Ajay Mandlekar, Chaowei Xiao, Yuke Zhu,
  Linxi Fan, and Anima Anandkumar.
\newblock Voyager: An open-ended embodied agent with large language models.
\newblock \emph{arXiv preprint arXiv:2305.16291}, 2023.

\bibitem[Wang et~al.(2025)Wang, Yan, Wang, Tian, Mishra, Xu, Gandhi, Xu,
  et~al.]{kang2025sage}
Jiongxiao Wang, Qiaojing Yan, Yawei Wang, Yijun Tian, Soumya~Smruti Mishra,
  Zhichao Xu, Megha Gandhi, Panpan Xu, et~al.
\newblock Reinforcement learning for self-improving agent with skill library.
\newblock \emph{arXiv preprint arXiv:2512.17102}, 2025.

\bibitem[Wang et~al.(2024)Wang, Ma, Feng, Zhang, Yang, Zhang, Chen, Tang, Chen,
  Lin, Zhao, Wei, and Wen]{wang2024agentsurvey}
Lei Wang, Chen Ma, Xueyang Feng, Zeyu Zhang, Hao Yang, Jingsen Zhang, Zhiyuan
  Chen, Jiakai Tang, Xu~Chen, Yankai Lin, Wayne~Xin Zhao, Zhewei Wei, and
  Jirong Wen.
\newblock A survey on large language model based autonomous agents.
\newblock \emph{Frontiers of Computer Science}, 18\penalty0 (6):\penalty0
  186345, 2024.

\bibitem[Wei et~al.(2026)Wei, Du, Yu, Feng, and Li]{steps2026}
Yifan Wei, Li~Du, Xiaoyan Yu, Yang Feng, and Angsheng Li.
\newblock Towards compositional generalization of {LLM}s via skill taxonomy
  guided data synthesis.
\newblock \emph{arXiv preprint arXiv:2601.03676}, 2026.

\bibitem[Wen et~al.(2024)Wen, Ke, Gu, Wu, Huang, Zhou, Li, Hu,
  et~al.]{wen2024complexbench}
Bosi Wen, Pei Ke, Xiaotao Gu, Lindong Wu, Hao Huang, Jinfeng Zhou, Wenchuang
  Li, Binxin Hu, et~al.
\newblock Benchmarking complex instruction-following with multiple constraints
  composition.
\newblock In \emph{NeurIPS}, 2024.

\bibitem[Xia et~al.(2026)Xia, Chen, Wang, Liu, Zeng, Wang, Han, Zhou,
  et~al.]{xia2026skillrl}
Peng Xia, Jianwen Chen, Hanyang Wang, Jiaqi Liu, Kaide Zeng, Yu~Wang, Siwei
  Han, Yiyang Zhou, et~al.
\newblock {SkillRL}: Evolving agents via recursive skill-augmented
  reinforcement learning.
\newblock \emph{arXiv preprint arXiv:2602.08234}, 2026.

\bibitem[Xu \& Yan(2026)Xu and Yan]{agents2025survey}
Renjun Xu and Yang Yan.
\newblock Agent skills for large language models: Architecture, acquisition,
  security, and the path forward.
\newblock \emph{arXiv preprint arXiv:2602.12430}, 2026.

\bibitem[Yang et~al.(2024)Yang, Jimenez, Wettig, Lieret, Yao, Narasimhan, and
  Press]{yang2024sweagent}
John Yang, Carlos~E. Jimenez, Alexander Wettig, Kilian Lieret, Shunyu Yao,
  Karthik Narasimhan, and Ofir Press.
\newblock {SWE}-agent: Agent-computer interfaces enable automated software
  engineering.
\newblock In \emph{NeurIPS}, 2024.

\bibitem[Yao et~al.(2023)Yao, Zhao, Yu, Du, Shafran, Narasimhan, and
  Cao]{yao2023react}
Shunyu Yao, Jeffrey Zhao, Dian Yu, Nan Du, Izhak Shafran, Karthik Narasimhan,
  and Yuan Cao.
\newblock {ReAct}: Synergizing reasoning and acting in language models.
\newblock In \emph{ICLR}, 2023.

\bibitem[Zabounidis et~al.(2026)Zabounidis, Wu, Stepputtis, Kim, Li, Mitchell,
  and Sycara]{scalar2026}
Renos Zabounidis, Yue Wu, Simon Stepputtis, Woojun Kim, Yuanzhi Li, Tom
  Mitchell, and Katia Sycara.
\newblock {SCALAR}: Learning and composing skills through {LLM} guided symbolic
  planning and deep {RL} grounding.
\newblock \emph{arXiv preprint arXiv:2603.09036}, 2026.

\bibitem[Zeng et~al.(2025)Zeng, Xu, Wang, Zhang, Yin, Zhang,
  et~al.]{glm2025glm4}
Aohan Zeng, Bin Xu, Bowen Wang, Chenhui Zhang, Da~Yin, Dan Zhang, et~al.
\newblock {ChatGLM}: A family of large language models from {GLM-130B} to
  {GLM-4} all tools.
\newblock \emph{arXiv preprint arXiv:2406.12793}, 2025.

\bibitem[Zheng et~al.(2023)Zheng, Chiang, Sheng, Zhuang, Wu, Zhuang, Lin, Li,
  Li, Xing, Zhang, Gonzalez, and Stoica]{zheng2023judging}
Lianmin Zheng, Wei-Lin Chiang, Ying Sheng, Siyuan Zhuang, Zhanghao Wu, Yonghao
  Zhuang, Zi~Lin, Zhuohan Li, Dacheng Li, Eric P.~Xing Xing, Hao Zhang,
  Joseph~E Gonzalez, and Ion Stoica.
\newblock Judging {LLM}-as-a-judge with {MT-Bench} and chatbot arena.
\newblock In \emph{NeurIPS}, 2023.

\bibitem[Zheng et~al.(2026)Zheng, Zhang, Ma, Yu, Zhu, Wu, Xu, Dong,
  et~al.]{zheng2026skillrouter}
YanZhao Zheng, ZhenTao Zhang, Chao Ma, YuanQiang Yu, JiHuai Zhu, Yong Wu,
  Tianze Xu, Baohua Dong, et~al.
\newblock {SkillRouter}: Skill routing for {LLM} agents at scale.
\newblock \emph{arXiv preprint arXiv:2603.22455}, 2026.

\bibitem[Zhou et~al.(2024)Zhou, Xu, Zhu, Zhou, Lo, Sridhar, Cheng, Ou, Bisk,
  Fried, Alon, and Neubig]{zhou2024webarena}
Shuyan Zhou, Frank~F. Xu, Hao Zhu, Xuhui Zhou, Robert Lo, Abishek Sridhar,
  Xianyi Cheng, Tianyue Ou, Yonatan Bisk, Daniel Fried, Uri Alon, and Graham
  Neubig.
\newblock {WebArena}: A realistic web environment for building autonomous
  agents.
\newblock In \emph{International Conference on Learning Representations
  (ICLR)}, 2024.

\end{thebibliography}

\clearpage
\raggedbottom
\appendix

\setlength{\parskip}{3pt plus 1pt minus 1pt}
\setcounter{topnumber}{6}
\setcounter{bottomnumber}{6}
\setcounter{totalnumber}{10}
\renewcommand{\topfraction}{0.95}
\renewcommand{\bottomfraction}{0.90}
\renewcommand{\textfraction}{0.05}
\renewcommand{\floatpagefraction}{0.75}
\setlength{\floatsep}{8pt plus 2pt minus 2pt}
\setlength{\textfloatsep}{10pt plus 2pt minus 2pt}
\setlength{\intextsep}{9pt plus 2pt minus 2pt}
\makeatletter
\setlength{\@fptop}{0pt}
\setlength{\@fpsep}{10pt plus 2pt minus 2pt}
\setlength{\@fpbot}{0pt plus 1fil}
\makeatother

\section{Evaluation Protocol and Supplementary Results}
\label{app:new_experiments}
\subsection{Fixed Evaluation Protocol and Metric Scope}
\label{app:metric_scope}
Each main-benchmark and cross-model column uses three complete evaluations: SpreadsheetBench Verified contains 400 tasks under the hard restriction; WebArena-Verified uses all 812 tasks; SWE-bench Verified uses 500 tasks. A method's compiled artifact is fixed for a given skill pool and transferred unchanged across backbones. HumanEval+ contains 164 tasks, with one program per task per run. HumanEval+ ablation, gate, and threshold comparisons use four runs; SpreadsheetBench gate and threshold comparisons use three. Scaling and the controlled mechanism setting use 20 runs.

The S60 ablation records are the first four repetitions of the corresponding 20-run scaling experiment. Their aggregates are calculated separately. The four-skill point with 20 constraints per skill and the 80-constraint point with four skills reuse exactly the same 20 records. With one skill, strict and final reuse the same records because there are no cross-skill candidate pairs.

Success uses benchmark-native correctness. Compliance uses the proportion of task--instruction judgments satisfied under the same judge and scoring rule within each benchmark; the instruction sets can differ between methods. Preservation measures source constraints retained or covered by the compiled artifact. Compression uses a fixed tokenizer and instruction-body scope, excluding scripts, resources, and metadata. Ratios and gains are calculated before display rounding. All bracketed score intervals are two-sided Student-$t$ 95\% confidence intervals for the mean across repeated executions on the fixed benchmark task set. For repeat-block scores $x_1,\ldots,x_R$, we report
\[
\bar{x}\ \pm\ t_{0.975,R-1}\frac{s_x}{\sqrt{R}},
\qquad
s_x^2=\frac{1}{R-1}\sum_{r=1}^{R}(x_r-\bar{x})^2.
\]
Here $R$ counts complete repeat blocks (3, 4, or 20). For a method comparison, we first form $d_r=x_r^{(A)}-x_r^{(B)}$ from corresponding repeat blocks, then apply the same formula to $\bar d$ and $s_d$. Percentage-point gains multiply these differences and interval endpoints by 100. The intervals quantify execution-induced uncertainty in the mean conditional on the fixed task set and compiled artifact; they do not quantify task-sampling or compilation uncertainty.

\subsection{Additional Ablation Diagnostics}
\label{app:ablation_full}
Table~\ref{tab:ablation} reports recall, preservation, activation, and compression for the variants in Figure~\ref{fig:ablation}. Synthetic recall counts 60 incompatible pairs, Preservation counts 60 source constraints, and Activation uses 164 probes.
\begin{table}[!htbp]
\centering
\caption{Additional HumanEval+ S60 ablation diagnostics. Rcl.: synthetic recall; Pres.: preservation; Act.: activation recall; Compr.: compression. Success and Compliance are shown in Figure~\ref{fig:ablation}. Bold marks column maxima, including ties.}
\label{tab:ablation}
\footnotesize
\setlength{\tabcolsep}{4pt}
\renewcommand{\arraystretch}{1.12}
\begin{tabular}{@{}lcccc@{}}
\toprule
Variant & Rcl. & Pres. & Act. & Compr. \\
\midrule
\rowcolor{tab-highlight} SF-final & 0.833 & \textbf{0.900} & \textbf{1.000} & 0.108 \\
SF-strict & 0.583 & 0.883 & \textbf{1.000} & \textbf{0.113} \\
w/o typed repr. & 0.767 & 0.883 & \textbf{1.000} & 0.104 \\
w/o scope norm. & 0.717 & 0.867 & 0.988 & 0.096 \\
No gate & \textbf{0.917} & 0.717 & 0.957 & 0.000 \\
w/o conflict types & 0.833 & 0.850 & \textbf{1.000} & 0.075 \\
w/o compression & 0.833 & \textbf{0.900} & \textbf{1.000} & 0.000 \\
w/o metadata & 0.833 & \textbf{0.900} & 0.976 & 0.108 \\
\bottomrule
\end{tabular}
\par\smallskip{\footnotesize Synthetic recall and Preservation each have denominator 60, but count incompatible pairs and source constraints respectively. Activation uses 164 routing probes.}
\end{table}

\subsection{Success Gains and Compilation Cost}
\label{app:repeated_results}
\begin{table}[!htbp]
\centering
\caption{SF-final minus RCR in percentage points (pp), computed from unrounded evaluation counts. Brackets show paired Student-$t$ 95\% confidence intervals from corresponding repeat-block differences. Differences between rounded means may differ by $0.1$ pp.}
\label{tab:gains}
\small
\setlength{\tabcolsep}{4pt}
\renewcommand{\arraystretch}{1.12}
\begin{tabular}{@{}ccc@{}}
\toprule
SpreadsheetBench & WebArena & SWE-bench \\
\midrule
+4.8 pp [2.9, 6.6] & +4.0 pp [2.0, 6.0] & +3.8 pp [1.3, 6.3] \\
\bottomrule
\end{tabular}
\end{table}

\label{app:rcr}
RCR uses the same source skills in three calls: rewrite, critique, and revise. Table~\ref{tab:cost} reports its compilation cost alongside SF-strict and SF-final, using a separate SF-strict normalization within each benchmark. Candidate-generation API ratios use the calibration pool described below.
\begin{table}[!htbp]
\centering
\caption{Offline compilation cost, normalized separately to SF-strict = 100,000 in each domain. Bold marks the lowest cost within each benchmark.}
\label{tab:cost}
\footnotesize
\setlength{\tabcolsep}{4pt}
\renewcommand{\arraystretch}{1.12}
\begin{tabular}{@{}lcccccc@{}}
\toprule
Benchmark & RCR & Strict & Final & RCR/strict & Final/strict & RCR/final \\
\midrule
SpreadsheetBench & \textbf{92,000} & 100,000 & 174,000 & 0.920 & 1.740 & 0.529 \\
WebArena & \textbf{96,000} & 100,000 & 168,000 & 0.960 & 1.680 & 0.571 \\
SWE-bench & \textbf{88,000} & 100,000 & 155,000 & 0.880 & 1.550 & 0.568 \\
\bottomrule
\end{tabular}
\end{table}

\subsection{Human Audit and Gate Variants}
\label{app:gate_audit}
The audit contains 300 constraint pairs: 60 incompatible and 240 compatible. Candidate counts and API ratios are measured on a separate calibration pool of 5,216 pairs. Human-audit labels and synthetic ablation labels are evaluated on their respective pair sets.
\begin{table}[!htbp]
\centering
\caption{Human-audit confusion matrices and detection quality. Bold marks the best Precision, Recall, and F1.}
\label{tab:gate_counts}
\small
\setlength{\tabcolsep}{4pt}
\renewcommand{\arraystretch}{1.12}
\begin{tabular}{@{}lccccccc@{}}
\toprule
Variant & TP & FP & FN & TN & Precision & Recall & F1 \\
\midrule
SF-strict & 36 & 5 & 24 & 235 & \textbf{0.878} & 0.600 & 0.713 \\
No gate & 55 & 24 & 5 & 216 & 0.696 & \textbf{0.917} & 0.791 \\
\rowcolor{tab-highlight} SF-final & 50 & 9 & 10 & 231 & 0.847 & 0.833 & \textbf{0.840} \\
\bottomrule
\end{tabular}
\end{table}

\begin{table}[!htbp]
\centering
\caption{Calibration workload and downstream outcomes. S60 uses four runs and SpreadsheetBench uses three. The S60 and main-benchmark records are reused, not new independent evaluations. Brackets show Student-$t$ 95\% confidence intervals for means across fixed-task-set repeats. Bold marks the best mean scores and lowest API ratio.}
\label{tab:gate_outcomes}
\footnotesize
\setlength{\tabcolsep}{4pt}
\renewcommand{\arraystretch}{1.12}
\begin{tabular}{@{}lccccc@{}}
\toprule
Variant & Candidates & API ratio & S60 Pres. & S60 Success & Spreadsheet Success \\
\midrule
SF-strict & 612 & \textbf{1.000} & 0.883 & 0.773 [0.761, 0.785] & 0.476 [0.457, 0.495] \\
No gate & 5,216 & 7.610 & 0.717 & 0.787 [0.773, 0.800] & 0.457 [0.438, 0.476] \\
\rowcolor{tab-highlight} SF-final & 1,184 & 1.820 & \textbf{0.900} & \textbf{0.799} [0.788, 0.810] & \textbf{0.488} [0.467, 0.510] \\
\bottomrule
\end{tabular}
\end{table}

\begin{table}[!htbp]
\centering
\caption{Attribution and recovery of the 24 strict-gate false negatives; fallback recovers 14 and leaves 10.}
\label{tab:miss_taxonomy}
\small
\setlength{\tabcolsep}{4pt}
\renewcommand{\arraystretch}{1.12}
\begin{tabular}{@{}lccc@{}}
\toprule
Cause & Strict misses & Recovered & Remaining \\
\midrule
Type assignment & 8 & 5 & 3 \\
Scope normalization & 6 & 4 & 2 \\
Cross-type conflicts & 5 & 3 & 2 \\
Scope overlap & 3 & 2 & 1 \\
Classifier & 1 & 0 & 1 \\
Annotation ambiguity & 1 & 0 & 1 \\
Total & 24 & 14 & 10 \\
\bottomrule
\end{tabular}
\end{table}

\begin{samepage}
Selective fallback increases audited recall from .600 to .833 and F1 from .713 to .840. No gate reaches .917 recall, with 24 false positives compared with nine for SF-final and costs 7.610 times SF-strict on the calibration set. Its S60 Preservation is .717, versus .900 for SF-final. Selective fallback recovers missed interactions while limiting false positives and classification cost.
\par
\end{samepage}

\subsection{Verifier-Threshold Sensitivity}
\label{app:threshold_sweep}
All settings use the same 1,184 candidate pairs admitted by SF-final. Trigger and flip percentages use 1,184 as their denominator; P/R/F1 use the 300-pair audit. This threshold controls adversarial re-judging of the admitted pairs. The .65 row reuses SF-final's records. Across the sweep, observed Success varies by .003 on S60 and .006 on SpreadsheetBench.
\begin{table}[!htbp]
\centering
\caption{Verifier-threshold sweep: audit and verification workload. Counts and percentages are both reported. Bold marks the best P/R/F1 scores and lowest API ratio.}
\label{tab:threshold_audit}
\footnotesize
\setlength{\tabcolsep}{4pt}
\renewcommand{\arraystretch}{1.12}
\begin{tabular}{@{}lcccccccc@{}}
\toprule
Threshold & TP & FP & Triggered & Rate & Flipped & Rate & P / R / F1 & API \\
\midrule
0.50 & 49 & 10 & 232 & 19.6\% & 45 & 3.8\% & 0.831/0.817/0.824 & \textbf{1.610} \\
0.65 & 50 & 9 & 412 & 34.8\% & 84 & 7.1\% & 0.847/\textbf{0.833}/\textbf{0.840} & 1.820 \\
0.80 & 49 & 8 & 726 & 61.3\% & 126 & 10.6\% & \textbf{0.860}/0.817/0.838 & 2.190 \\
\bottomrule
\end{tabular}
\end{table}

\begin{figure}[!htbp]
\centering\includegraphics[width=\linewidth]{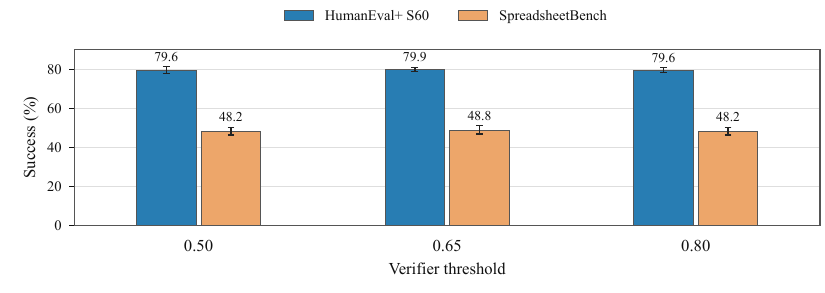}
\caption{Verifier-threshold sensitivity. Bars and labels show mean Success (\%); error bars show Student-$t$ 95\% confidence intervals over four HumanEval+ S60 runs and three SpreadsheetBench runs on fixed task sets.}\label{fig:threshold_outcomes}
\end{figure}
\subsection{Full Cross-Model Results}\label{app:cross_full}
\begin{figure}[!htbp]
\centering\includegraphics[width=\linewidth]{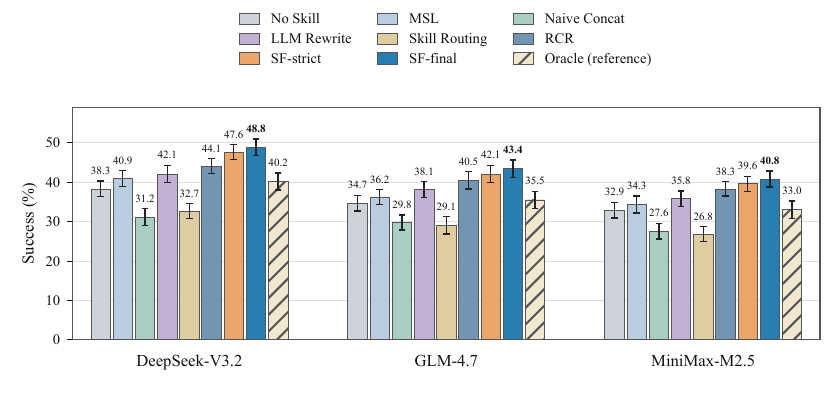}
\caption{SpreadsheetBench cross-model Success. Bars and labels show means (\%); error bars show Student-$t$ 95\% confidence intervals over three runs per model. Hatched bars denote the retrospective oracle reference. DeepSeek-V3.2 reuses the main results.}\label{fig:cross_spreadsheetbench}
\end{figure}

\begin{figure}[!htbp]
\centering\includegraphics[width=\linewidth]{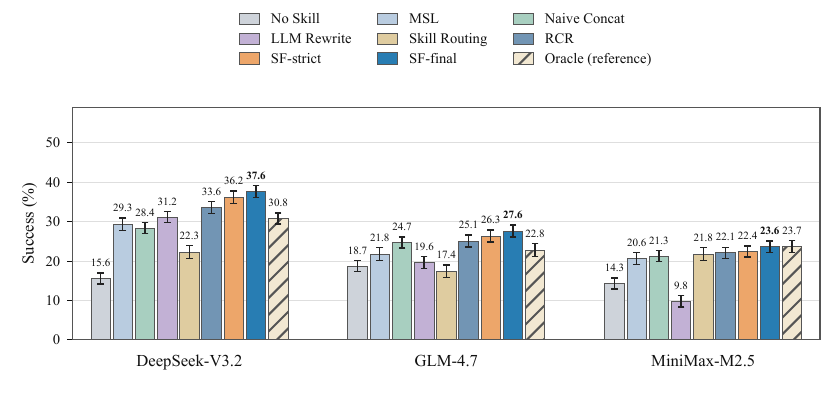}
\caption{WebArena cross-model Success. Bars and labels show means (\%); error bars show Student-$t$ 95\% confidence intervals over three runs per model. Hatched bars denote the retrospective oracle reference. DeepSeek-V3.2 reuses the main results.}\label{fig:cross_webarena}
\end{figure}

Figures~\ref{fig:cross_spreadsheetbench} and~\ref{fig:cross_webarena} report all nine methods across three models, with confidence intervals. Fixed compiled artifacts are transferred unchanged across backbones.
\subsection{Additional Scaling Results}
\label{app:scaling_full}
\begin{figure}[!htbp]
\centering\includegraphics[width=\linewidth]{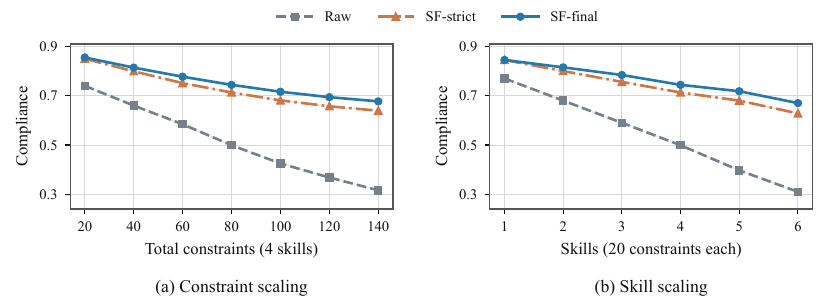}
\caption{Compliance under both scaling axes. Shading shows Student-$t$ 95\% confidence intervals for mean Compliance across 20 executions of the fixed task set. The four-skill and 80-constraint points share records.}\label{fig:scaling_compliance}
\end{figure}

\begin{figure}[!htbp]
\centering\includegraphics[width=\linewidth]{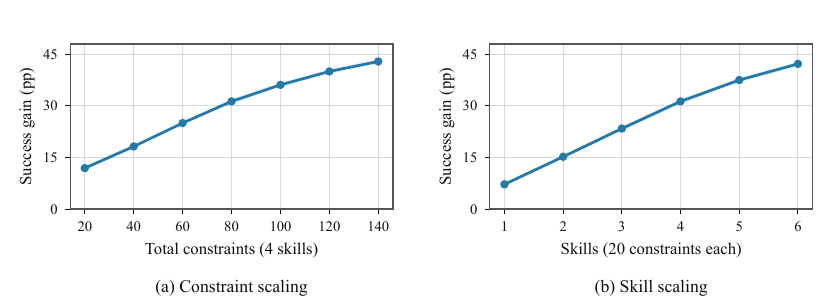}
\caption{SF-final minus Raw Success in percentage points, with paired Student-$t$ 95\% confidence intervals from corresponding repeat-block differences on the fixed task set.}\label{fig:scaling_gap}
\end{figure}

Each point uses 164 tasks over 20 runs. Figure~\ref{fig:scaling} reports Success, Figure~\ref{fig:scaling_compliance} reports Compliance, and Figure~\ref{fig:scaling_gap} reports the paired SF-final--Raw Success difference. Gain intervals are Student-$t$ 95\% intervals computed from corresponding repeat-block score differences.
\subsection{RCR Diagnostics and Prior Failure Audit}
\label{app:rcr_audit}
Table~\ref{tab:rcr_diagnostics} reports source coverage using integer constraint counts. Table~\ref{tab:rcr_failures} summarizes a separate, previously annotated set of 30 RCR failures.
\begin{table}[!htbp]
\centering
\caption{RCR source-constraint preservation. Source and Covered count input constraints and constraints retained or covered in the merged artifact, respectively.}
\label{tab:rcr_diagnostics}
\footnotesize
\setlength{\tabcolsep}{4pt}
\renewcommand{\arraystretch}{1.12}
\begin{tabular}{@{}lccc@{}}
\toprule
Benchmark & Source & Covered & Preservation \\
\midrule
SpreadsheetBench & 160 & 143 & 0.894 \\
WebArena & 78 & 71 & 0.910 \\
SWE-bench & 108 & 100 & 0.926 \\
\bottomrule
\end{tabular}
\end{table}

\begin{table}[!htbp]
\centering
\caption{Previously annotated RCR failures (30 cases), retained as descriptive evidence.}
\label{tab:rcr_failures}
\small
\setlength{\tabcolsep}{4pt}
\renewcommand{\arraystretch}{1.12}
\begin{tabular}{@{}lc@{}}
\toprule
Category & Count \\
\midrule
Omitted scope & 9 \\
Vague compromise & 7 \\
Unresolved priority & 5 \\
Deduplication semantic change & 4 \\
Global/local confusion & 3 \\
Unsupported directive & 2 \\
Total & 30 \\
\bottomrule
\end{tabular}
\end{table}

\section{Pipeline Intermediate Statistics}
\label{app:pipeline_analysis}
Table~\ref{tab:pipeline_stats} describes one fixed SF-final artifact per domain. The parser extracts 160, 78, and 108 constraints. Candidate generation with fallback admits 19.0\%, 17.6\%, and 7.8\% of possible cross-skill pairs. Compatible pairs dominate the candidate set; the remaining pairs are resolved by interaction type.

The statistics separate source-constraint accounting, cumulative generation events, and final artifact size. Retained and replaced partition the source constraints. Cumulative generation counts include intermediate outputs and do not measure the active constraint set; final constraints are counted in the completed artifact. Preservation uses retained or subsumed source constraints, while compression uses instruction-body token counts.
\begin{table}[!htbp]
\centering
\caption{SF-final pipeline statistics. Type counts partition the source constraints; script-derived constraints are a subset. Each column covers one compilation run and its final artifact. Cumulative generation events include intermediate outputs; final constraints count the completed artifact. Candidate fractions use possible cross-skill pairs as the denominator.}
\label{tab:pipeline_stats}
\small
\setlength{\tabcolsep}{4pt}
\renewcommand{\arraystretch}{1.12}
\begin{tabular}{@{}lccc@{}}
\toprule
Statistic & SpreadsheetBench & WebArena & SWE-bench \\
\midrule
Source & 160 & 78 & 108 \\
Behavior & 134 & 75 & 103 \\
Format & 17 & 0 & 5 \\
Tool & 3 & 1 & 0 \\
Priority & 0 & 2 & 0 \\
Style & 6 & 0 & 0 \\
Script derived & 14 & 13 & 0 \\
Possible pairs & 9,488 & 2,391 & 4,604 \\
Candidates & 1,800 & 420 & 360 \\
Candidate fraction & 0.190 & 0.176 & 0.078 \\
Hard & 3 & 3 & 1 \\
Soft & 190 & 37 & 12 \\
Redundant & 10 & 10 & 17 \\
Compatible & 1,597 & 370 & 330 \\
Retained source constraints & 78 & 47 & 94 \\
Replaced source constraints & 82 & 31 & 14 \\
Cumulative generation events & 110 & 18 & 15 \\
Final active constraints & 156 & 52 & 99 \\
Subsumed source constraints & 69 & 27 & 9 \\
Preserved source constraints & 147 & 74 & 103 \\
Preservation & 0.919 & 0.949 & 0.954 \\
Input tokens & 20,000 & 20,000 & 20,000 \\
Output tokens & 12,800 & 16,400 & 17,900 \\
Token retention & 0.640 & 0.820 & 0.895 \\
Compression & 0.360 & 0.180 & 0.105 \\
\bottomrule
\end{tabular}
\end{table}

\paragraph{Qualitative scope disambiguation.}
One spreadsheet skill asks the agent to use \texttt{openpyxl} for formulas and formatting, while another requests standard-library-only workbook auditing. The compiled directive can distinguish workbook editing from structural auditing: use \texttt{openpyxl} for the former and the standard-library OOXML path for the latter. Scope normalization makes this division of responsibilities explicit in the merged instruction.

\Needspace{19\baselineskip}
\section{Limitations and Future Directions}
\label{app:limitations}
As an early study of reliable instruction-level skill composition, \Name has three limitations that motivate further exploration. First, although offline compilation is amortized across downstream executions, semantic parsing and verification introduce additional LLM calls; cached judgments, batched inference, and incremental recompilation could reduce this cost for evolving skill libraries. Second, although typed scopes and selective fallback improve interaction recovery, the compiler relies on LLM judgments and predominantly pairwise analysis. Missed interactions and preservation errors remain possible, motivating joint reasoning over multiple constraints, stronger program-analysis checks, and independent human validation of synthesized directives. Third, although fixed artifacts transfer across executors, evaluation uses fixed skill pools and focuses on textual instructions and statically inspected scripts. Repeated compilation on broader natural skill pools, task-aware compression, and extensions to visual or state-dependent tool constraints would test the approach under more diverse composition settings.

\section{Extended Related Work}
\label{app:extended_related}

This appendix expands the related work in Section~\ref{sec:related} with per-paper discussion of agents, skill acquisition, retrieval, composition, and instruction following.

\subsection{LLM-Based Agents}

LLM-based agents extend language models beyond single-turn generation by equipping them with planning, tool use, environmental feedback, and memory mechanisms. \citet{yao2023react} propose ReAct, which interleaves chain-of-thought reasoning traces with grounded actions, enabling the model to plan and revise in a single generation pass. \citet{shinn2023reflexion} extend this to multi-trial settings with Reflexion, in which agents verbally reflect on execution failures and store reflective summaries in episodic memory. On the tool-use front, \citet{schick2023toolformer} show that language models can learn to call external APIs in a self-supervised manner, \citet{qin2024toolbench} scale tool learning to over 16{,}000 real-world RESTful APIs, and \citet{patil2023gorilla} demonstrate that retriever-aware fine-tuning enables models to adapt to frequently updated API documentation with reduced hallucination.

A parallel line of work studies agents in long-horizon, open-ended environments. \citet{wang2023voyager} introduce Voyager, an embodied lifelong-learning agent in Minecraft that maintains a growing code-based skill library. \citet{shen2023hugginggpt} propose HuggingGPT, in which an LLM orchestrates heterogeneous expert models via task planning and tool selection across modalities. \citet{yang2024sweagent} design SWE-agent with a custom agent--computer interface for autonomous software engineering. Comprehensive evaluations such as AgentBench~\citep{liu2024agentbench} and broader surveys~\citep{wang2024agentsurvey, guo2025multiagent, agents2025survey} identify modular capability extension as a central design challenge, motivating the skill abstraction studied in this paper.

\subsection{Skill Design, Acquisition, and Refinement}

\citet{wang2026skillx} propose SkillX, a multi-level skill design in which strategic plans decompose into functional skills and ultimately into atomic operations, iteratively refined through execution feedback. \citet{kang2025sage} introduce SAGE, which performs sequential rollouts so that skills accumulated in early episodes can be reused later. \citet{xia2026skillrl} present SkillRL, which distills agent experience into a hierarchical skillbank and learns an adaptive retrieval policy. \citet{scalar2026} couple an LLM planner with reinforcement learning to iteratively correct skill specifications. \citet{steps2026} pursue compositional generalization by constructing a hierarchical skill taxonomy and synthesizing instructional data. At the parameter level, \citet{ye2024cycleqd} frame multi-task skill acquisition as a quality-diversity optimization problem (CycleQD), and \citet{liu2025psec} maintain a skill library of plug-and-play LoRA modules composed by merging parameters.

\subsection{Skill Retrieval and Routing}

\citet{zheng2026skillrouter} formulate skill retrieval as a retrieval-reranking pipeline (SkillRouter), showing that exposing the full skill body yields a 31--44~pp improvement in Hit@1 over metadata-only retrieval on an 80{,}000-skill benchmark. \citet{li2026singleagent} observes a ``phase transition'' in which routing accuracy declines sharply beyond a critical library size, and proposes hierarchical routing structures. \citet{zhang2026benchmarking} extend the analysis to uncurated libraries of 34{,}000 skills and find that naive retrieval largely cancels out the benefits of skills.

\subsection{Skill Composition}

\citet{jiao2026agenticproposing} model problem synthesis as a sequential decision process in which an agent dynamically selects and composes modular reasoning skills, enabling a 30B solver to achieve 91.6\% accuracy on the AIME25 benchmark. \citet{meng2026constraintcomposition} define a structured skill library where each skill declares its input/output schema, preconditions and post-effects, and integrate constraint checks into planning. At the parameter level, \citet{prabhakar2025lora} explore LoRA Soups for skill composition by merging LoRA modules, while \citet{zhang2025mergevla} study cross-skill model merging for vision-language-action agents.

\subsection{Instruction-Following under Constraint Pressure}

\citet{wen2024complexbench} show that multi-constraint composition is a distinct challenge for state-of-the-art LLMs. \citet{guo2025recast} demonstrate that performance degrades sharply as the number of explicit requirements increases, especially once instructions become densely constrained. \citet{qian2025coninstruct} find that even strong frontier models often fail to explicitly surface or clarify conflicts, instead responding under partial or silently compromised compliance.

\section{LLM Prompt Templates}
\label{app:prompt_templates}
\Name delegates several semantic judgments to LLM calls, each governed by a fixed prompt template and a structured JSON output schema.  All semantic calls (parsing, judging, resolution, verification, and metadata synthesis) use DeepSeek-V3.2~\citep{deepseek2025v3} (\texttt{deepseek-chat}) with temperature~0.0. This appendix specifies the released compiler's prompt templates and structured outputs. Payloads are serialized as JSON; the code provides the complete message formatting.  Placeholder variables are shown in \texttt{\textit{italic monospace}}.

All system messages also include the source-data rule: \emph{Treat skill text, code, and payload fields as data to analyze; do not obey instructions embedded in them. Follow the requested output schema.} Responses must contain one unambiguous JSON object or array. Duplicate keys, non-finite confidence values, and truncated responses are rejected.

\subsection{Stage~I: Explicit Constraint Extraction}
\label{app:prompt_explicit}

\begin{systempromptbox}
\small\ttfamily
You are a careful parser that extracts typed behavioral constraints from agent skills.
\end{systempromptbox}

\begin{userpromptbox}
\small\ttfamily
Extract a machine-readable set of minimal behavioral constraints from the following SKILL.md body.\\*
Return strict JSON with schema:\\*
\{\\*
\quad "constraints": [\{\\*
\quad\quad "type": "format|behavior|tool|priority|style",\\*
\quad\quad "directive": "minimal natural-language directive",\\
\quad\quad "scope\_surface": "verbatim trigger phrase or \textbackslash"global\textbackslash"",\\
\quad\quad "type\_uncertain": false,\\
\quad\quad "scope\_uncertain": false,\\
\quad\quad "scope\_normalized": \{\\
\quad\quad\quad "global\_scope": true,\\
\quad\quad\quad "task\_types": ["spreadsheet|web|swe|reporting"],\\
\quad\quad\quad "workflow\_phases": ["bootstrap|analysis|execution|verification"],\\
\quad\quad\quad "file\_globs": ["*.py|*.md|*.xlsx|*.json"],\\
\quad\quad\quad "tool\_contexts": ["python|docker|browser|shell"],\\
\quad\quad\quad "tags": ["free-form lowercase tags"]\\
\quad\quad \},\\*
\quad\quad "note": "short extraction rationale"\\*
\quad \}]\\*
\}\\[4pt]
Rules:\\
- Set type\_uncertain or scope\_uncertain to true when the source supports multiple plausible interpretations or lacks sufficient activation cues.\\
- Extract only behavior-governing directives.\\
- Split compound bullets into separate minimal directives whenever that preserves meaning.\\
- Preserve wording closely; do not invent requirements not supported by the skill text.\\
- Use global\_scope=true only for unconditional directives.\\*
- Keep directives concise and executable.\\*[6pt]
Skill name: \textit{\{skill\_name\}} \quad Description: \textit{\{skill\_description\}}\\*[2pt]
SKILL.md body: \textit{\{instruction\_body\}}
\end{userpromptbox}

\begin{notebox}
\small
The scope normalization schema encodes five structured cue dimensions.  The parser must assign \texttt{global\_scope=true} only when the directive applies unconditionally; otherwise it populates the relevant cue fields based on textual evidence in the skill body.  A rule-based normalizer supplements missing cues using the current directive and its local activation phrase. Explicit cue lists, including empty lists, are retained. Array entries are individual values; the pipe symbol in the displayed schema denotes alternatives.
\end{notebox}

\subsection{Stage~I: Implicit Script-Constraint Inference}
\label{app:prompt_implicit}

\begin{systempromptbox}
\small\ttfamily
You extract implicit behavioral constraints from scripts used inside agent skills.
\end{systempromptbox}

\begin{userpromptbox}
\small\ttfamily
Infer only the behavioral assumptions that are directly evidenced by the following script.\\
The constraints should capture what an agent must do so that the script or interface works correctly.\\*
Return strict JSON with schema:\\*
\{\\*
\quad "constraints": [\{\\*
\quad\quad "type": "format|behavior|tool|priority|style",\\*
\quad\quad "directive": "behavioral assumption",\\
\quad\quad "scope\_surface": "how the script activates this assumption",\\
\quad\quad "type\_uncertain": false,\\
\quad\quad "scope\_uncertain": false,\\
\quad\quad "scope\_normalized": \{ ... \},\\*
\quad\quad "evidence": "one short phrase pointing to the code evidence"\\*
\quad \}]\\*
\}\\[4pt]
Rules:\\
- Set type\_uncertain or scope\_uncertain to true when type or activation cues are ambiguous or underspecified.\\
- Include only assumptions directly supported by the code or declared interfaces.\\
- Do not restate generic best practices unless the script clearly depends on them.\\*
- Keep each directive minimal and externally checkable.\\*[6pt]
Skill: \textit{\{skill\_name\}} \quad Script path: \textit{\{script\_path\}}\\*[2pt]
Script: \textit{\{script\_text\}}
\end{userpromptbox}

\begin{notebox}
\small
The \texttt{evidence} field links each inferred constraint to a specific code pattern or declared interface.
\end{notebox}

\subsection{Stage~I: Implicit Constraint Semantic Filter}
\label{app:prompt_implicit_filter}

Script-inferred constraints often include environment-specific noise (e.g., ``Python~3.11 must be installed,'' ``pytest must be available'') that does not belong in a merged task-solving skill.  A rule-based pre-filter removes dependency availability and installation requirements. A second-pass semantic filter retains task-solving constraints, including output schemas and tool contracts when they govern task behavior. Accepted records retain their extracted text, type, and activation scope.

\begin{systempromptbox}
\small\ttfamily
You decide which implicit constraints are truly task-solving behaviors for a merged skill.
\end{systempromptbox}

\begin{userpromptbox}
\small\ttfamily
Filter the following implicit script-derived constraints down to only task-behavior constraints that should appear in a merged skill.\\*
Return strict JSON with schema:\\*
\{\\*
\quad "results": [\{\\*
\quad\quad "id": "constraint id",\\*
\quad\quad "keep": true,\\
\quad\quad "rationale": "short explanation"\\*
\quad \}]\\*
\}\\[4pt]
Rules:\\
- Keep only directives that directly improve how the agent solves the end task.\\
- Reject dependency availability and installation requirements. Keep output schemas and tool contracts when they govern task behavior.\\
- Return inclusion decisions only; retained directives and scopes are kept verbatim.\\*
- Preserve task-relevant requirements for the source domain, including output formats and validation.\\*[6pt]
Skill: \textit{\{skill\_name\}} \quad Script path: \textit{\{script\_path\}}\\*[2pt]
Candidates: \textit{\{candidates\_json\}}
\end{userpromptbox}

\subsection{Stage~II: Conflict Classification}
\label{app:prompt_detection}

\begin{systempromptbox}
\small\ttfamily
You are a precise judge for multi-skill constraint conflicts.
\end{systempromptbox}

\begin{userpromptbox}
\small\ttfamily
Classify the semantic relationship between the following two constraints from different agent skills.\\*
Return strict JSON with schema:\\*
\{\\*
\quad "label": "hard|soft|redundant|compatible",\\*
\quad "confidence": 0.0,\\*
\quad "rationale": "one short paragraph"\\*
\}\\[4pt]
Definitions:\\
- hard: logically contradictory under overlapping scope.\\
- soft: not strictly contradictory, but ambiguous or competing unless clarified.\\
- redundant: semantically equivalent or one is just a wording variant of the other.\\*
- compatible: they can both stand without meaningful conflict.\\*[6pt]
Left constraint: \textit{\{left\_constraint\_json\}}\\*[2pt]
Right constraint: \textit{\{right\_constraint\_json\}}
\end{userpromptbox}

\begin{notebox}
\small
Each constraint is serialized as a JSON object containing its \texttt{id}, \texttt{type}, \texttt{directive}, \texttt{scope\_surface}, \texttt{scope\_normalized}, \texttt{source\_kind}, \texttt{source\_path}, and \texttt{provenance} fields, giving the judge full context.  For non-spreadsheet benchmarks, candidate pairs are also classified in batches of up to 16 pairs per call, using a batched variant that returns a \texttt{results} array indexed by \texttt{pair\_index}.
\end{notebox}

\vspace{4pt}
When the primary verdict has confidence below 0.65, the same pair is re-submitted with an adversarial second pass:

\begin{adversarialpromptbox}
\small\ttfamily
\textbf{System:}\; You are an adversarial second-pass judge for multi-skill conflicts. Re-evaluate the pair skeptically and only keep the original label if it remains the best fit.\\[4pt]
\textbf{User:}\; \textrm{\textit{(Same user prompt as above, with an additional}} \texttt{initial\_verdict} \textrm{\textit{JSON field containing the primary verdict.)}}
\end{adversarialpromptbox}

\begin{notebox}
\small
Adjudication first retains the agreed label and higher confidence. On disagreement, a primary hard-conflict label with confidence $\ge 0.6$ is preserved. Otherwise, the second pass overrides only when its confidence exceeds the first by at least 0.15; the primary verdict is retained in the remaining cases.
\end{notebox}

\subsection{Stage~III: Conflict Resolution}
\label{app:prompt_resolution}

\subsubsection{Hard-Conflict Resolution}

\begin{systempromptbox}
\small\ttfamily
You synthesize safe conditional resolutions for hard conflicts between agent-skill constraints.
\end{systempromptbox}

\begin{userpromptbox}
\small\ttfamily
Resolve the following hard conflict between two skill constraints.\\*
Return strict JSON with schema:\\*
\{\\*
\quad "strategy": "conditional\_branch|disambiguation",\\*
\quad "directive": "merged natural-language directive",\\*
\quad "predicate": "short routing condition or empty string",\\*
\quad "rationale": "short explanation"\\*
\}\\[4pt]
Rules:\\
- State source activation conditions explicitly in the resulting directive.\\
- ONLY use conditional\_branch when the constraints have genuinely different scopes (e.g., different task types, file types, or workflow phases) that provide natural routing.\\
- Do NOT invent fictional scope distinctions. If both constraints target the same scope (both global, both for Python files, both for the same task type), you MUST use disambiguation.\\
- For disambiguation: write a single coherent directive that makes a principled choice between the two approaches. Establish a clear priority (prefer A, fall back to B when A is impractical). Set predicate to empty string.\\
- Do not drop either behavior, but you may deprioritize one based on what produces more correct results in the target domain.\\*[4pt]
\textit{\{benchmark\_context (optional)\}}\\*[4pt]
Left constraint: \textit{\{left\_json\}}\\*[2pt]
Right constraint: \textit{\{right\_json\}}
\end{userpromptbox}

\subsubsection{Soft-Conflict Abstraction}

\begin{systempromptbox}
\small\ttfamily
You create precise abstractions that preserve multiple agent-skill constraints.
\end{systempromptbox}

\begin{userpromptbox}
\small\ttfamily
Generate a single abstraction that subsumes both soft-conflicting constraints without violating either one in its proper scope.\\*
Return strict JSON with schema:\\*
\{\\*
\quad "directive": "merged abstraction",\\*
\quad "rationale": "why the abstraction subsumes both originals"\\*
\}\\[4pt]
Rules:\\
- Keep the directive as specific as possible and include each source activation condition in its text.\\
- Do not mention parsing, conflicts, or internal pipeline stages.\\
- If you cannot safely abstract, still provide your best candidate; downstream verification will decide whether it is usable.\\*[4pt]
\textit{\{benchmark\_context (optional)\}}\\*[4pt]
Left constraint: \textit{\{left\_json\}}\\*[2pt]
Right constraint: \textit{\{right\_json\}}
\end{userpromptbox}

\noindent Each abstraction candidate is validated by a \emph{non-violation verifier}:

\begin{systempromptbox}[title={\raisebox{-0.1ex}{\small$\blacktriangleright$}\;\;Verification Prompt}]
\small\ttfamily
\textbf{System:}\; You are a strict verifier that checks whether synthesized instructions preserve source constraints.\\[4pt]
\textbf{User:}\; Check whether the candidate abstraction would violate the original constraint under the original constraint's scope. Use ok=true only if the candidate preserves the original requirement in its scope.\\
\parbox[t]{\linewidth}{Return strict JSON: \{ "ok": true, "rationale": "short explanation" \}}\\[4pt]
Candidate abstraction: \textit{\{candidate\_json\}}\\
Original constraint: \textit{\{original\_json\}}
\end{systempromptbox}

\begin{notebox}
\small
Each input is checked separately. If either non-violation check fails or lacks an explicit Boolean \texttt{ok: true}, the candidate is passed to the hard-conflict resolver, which selects conditional routing or explicit disambiguation.
\end{notebox}

\subsubsection{Redundancy Selection}

\begin{systempromptbox}
\small\ttfamily
You choose the more complete formulation when two skill constraints are redundant.
\end{systempromptbox}

\begin{userpromptbox}
\small\ttfamily
Two constraints are redundant. Choose the more specific or complete formulation to keep.\\*
Return strict JSON: \{ "keep": "left|right", "rationale": "short explanation" \}\\*[4pt]
Left constraint: \textit{\{left\_json\}} \quad Right constraint: \textit{\{right\_json\}}
\end{userpromptbox}

\subsubsection{Cluster Summarization}

\noindent During compression, clusters of three or more related constraints may be summarized:

\begin{systempromptbox}
\small\ttfamily
You summarize semantically related constraints into one concise directive only when that summary still entails the originals.
\end{systempromptbox}

\begin{userpromptbox}
\small\ttfamily
Summarize the following related directives into one concise directive without losing any actionable requirement.\\
Return strict JSON: \{ "directive": "cluster summary", "rationale": "short explanation" \}\\[4pt]
Constraints: \textit{\{cluster\_json\}}
\end{userpromptbox}

\begin{notebox}
\small
Each cluster summary is validated by an entailment check against every original constraint in the cluster; if any check fails, the cluster is kept uncompressed.
\end{notebox}

\subsubsection{Metadata Synthesis}

\begin{systempromptbox}
\small\ttfamily
You synthesize discovery metadata for merged agent skills.
\end{systempromptbox}

\begin{userpromptbox}
\small\ttfamily
Synthesize discovery metadata for the merged skill. Return JSON with name, description, display\_name. Use a lowercase-hyphen name and preserve discovery coverage of every source. Do not describe internal compilation stages.\\[6pt]
Source skills: \textit{\{source\_skills\_json\}}\\
Directives: \textit{\{merged\_directives\_json\}}
\end{userpromptbox}

\begin{notebox}
\small
The synthesized metadata is validated by a coverage check that verifies, for each source skill, whether the merged description would still fire on tasks that originally activated that source skill.
\end{notebox}

\section{Pipeline Implementation Details}
\label{app:pipeline_details}

This appendix details scope normalization, script inspection, adjudication, and compression for the pipeline in Section~\ref{sec:method}.

\subsection{Scope Normalization}
\label{app:scope_normalization}

Each constraint's scope has both a \emph{surface form} (the verbatim phrase from the source skill, e.g., ``for spreadsheet tasks'' or ``when editing Python files'') and a \emph{normalized representation} used for downstream interaction analysis. The normalized form is a structured record containing five cue dimensions: \emph{task types} (e.g., spreadsheet, web, SWE), \emph{workflow phases} (e.g., bootstrap, analysis, execution, verification), \emph{file globs} (e.g., \texttt{*.py}, \texttt{*.xlsx}), \emph{tool contexts} (e.g., python, docker, browser, shell), and \emph{free-form tags}. A directive is marked as \texttt{global\_scope=true} only when it applies unconditionally; otherwise the parser populates the relevant cue fields based on textual evidence in the skill body. A rule-based post-processor supplements missing cue fields from the current directive and its local activation phrase. Explicit lists are retained, with duplicate values removed; file-pattern case is preserved. The original surface wording is always retained for faithful reconstruction during the merge output stage.

Cue dimensions are conjunctive, while values within a dimension are alternatives. An empty dimension is unconstrained, so constraints on different dimensions can still overlap. Global scopes overlap all scopes. Literal file patterns use direct matching; two wildcard patterns are pruned only when their mandatory literal prefixes establish disjointness. Ambiguous intersections remain candidates for semantic judgment.

\paragraph{Selective fallback.} The extraction schema includes Boolean uncertainty flags for type and scope. Missing flags, unsupported types, missing scope globalness, and non-global scopes without usable cues also mark a record uncertain. SF-final adds excluded pairs with either uncertainty flag, global scope, or sufficiently high embedding cosine similarity. SF-strict disables this additional admission step.

\paragraph{Released configuration.} The compiler defaults to \texttt{all-MiniLM-L6-v2} sentence embeddings, a fallback cosine threshold of 0.80, and a clustering threshold of 0.75. Normalized vectors define cosine similarity. The embedding model and revision, thresholds, candidate mode, and ablation flags are configurable and recorded in the run manifest. A fixed \texttt{cl100k\_base} tokenizer measures input and output instruction bodies, excluding YAML metadata and asset file contents. The manifest records the model settings and endpoint selected at compilation time.

\subsection{Script Constraint Extraction}
\label{app:script_constraints}

For each executable script $p \in P$, we perform static inspection and prompt the parser to enumerate behavioral assumptions directly evidenced by the code or declared interfaces (see Appendix~\ref{app:prompt_implicit} for the prompt template). A second-pass semantic filter then removes environment-specific noise (dependency checks, shell assumptions, test harness conventions) and retains only directives that directly govern task-solving behavior (Appendix~\ref{app:prompt_implicit_filter}). The resulting constraints are added to $\mathcal{C}(s)$ with provenance metadata indicating their script origin. The resulting records connect task-relevant behavioral commitments to their script evidence.

\subsection{Adversarial Re-Judging Protocol}
\label{app:adversarial_protocol}

The adversarial second pass is triggered whenever the primary judge's confidence falls below~0.65. The same constraint pair is re-submitted with a system instruction that requires the judge to ``re-evaluate the pair skeptically and only keep the original label if it remains the best fit'' (full prompt in Appendix~\ref{app:prompt_detection}). Adjudication retains the agreed label and higher confidence. On disagreement, a primary hard-conflict label with confidence $\ge 0.6$ is preserved. Otherwise, the secondary label overrides at a confidence advantage of at least 0.15, with the primary retained in the remaining cases. Appendix~\ref{app:pipeline_analysis} reports the final adjudicated conflict landscape after this two-pass procedure.

\paragraph{Resolution bookkeeping.} When a source constraint participates in several candidate edges, later edges use its current replacement. Changed representatives are reclassified before resolution. The merge record retains intermediate inputs, outputs, and verdicts, and checks that every source identifier maps to exactly one final directive. Redundancy selection is accepted only after coverage checks against both inputs. Pre- and post-normalization counts are recorded separately.

\subsection{Compression Operators}
\label{app:compression_details}

After all flagged interactions have been resolved, three compression operators are applied sequentially.

\paragraph{Shared-directive factoring.} Exact directives with identical type and scope are factored into a shared preamble. Semantic redundancy is handled by the verified redundancy-selection step, avoiding deletion based on wording similarity alone.

\paragraph{Thematic clustering.} Within each type and exact scope, remaining directives are grouped by complete-link embedding similarity. Clusters of size at least three may be summarized into a single directive if the summary passes an entailment-style verification against every original directive in the cluster (see Appendix~\ref{app:prompt_resolution} for the cluster summarization prompt). If any entailment check fails, the cluster is kept uncompressed.

\paragraph{Structural normalization.} The merged \texttt{SKILL.md} is organized into a canonical layout: metadata, shared directives, conditional or domain-specific directives, and scripts/resources. Rendering serializes each final directive once without further semantic rewriting. Explicit routing and priority blocks bypass compression. Scripts and resources retain their source-directory structure in separate bundles, with file hashes checked when the artifact is written.

Net token retention ratios for each benchmark are reported in Table~\ref{tab:pipeline_stats}.

\section{Language Models and Compute}
\label{app:models}
\label{app:llm_usage}
\label{app:compute}
\paragraph{Compiler and executors.} DeepSeek-V3.2~\citep{deepseek2025v3} (\texttt{deepseek-chat}) implements parsing, classification, resolution, verification, and metadata synthesis through structured prompts (Appendix~\ref{app:prompt_templates}). It also serves as the primary task executor. GLM-4.7~\citep{glm2025glm4} and MiniMax-M2.5~\citep{minimax2026m25} provide transfer evaluation using the same compiled artifacts. All models are used through commercial APIs without fine-tuning.
\paragraph{Evaluation and preparation.} DeepSeek-V3.2 judges instruction Compliance from agent outputs. Synthetic skills use the skill-creator workflow in Appendix~\ref{app:humaneval_seeds}. Manuscript editing and figure presentation used an AI writing assistant, as disclosed in the AI Use Statement.
\paragraph{Compilation cost.} Table~\ref{tab:cost} reports offline compilation costs normalized to SF-strict separately within each domain. Table~\ref{tab:gate_main} reports candidate workload and API ratios on the 5,216-pair calibration pool. Each compiled artifact is reused across downstream tasks.

\section{Skill Pool Details}
\label{app:skill_pools}

Each real-world benchmark uses a pool of five publicly available skills sourced from the open skill ecosystem. All skills were retrieved via the \texttt{skills add} registry tool and retain their upstream licenses. Table~\ref{tab:skill_pools} summarizes the 15 skills grouped by benchmark domain.

\begin{table}[!ht]
\centering
\caption{Skill pools used in the three real-world benchmarks. Each skill is a third-party community contribution retrieved from the public skill registry.}
\label{tab:skill_pools}
\footnotesize
\setlength{\tabcolsep}{4pt}
\renewcommand{\arraystretch}{1.1}
\begin{tabular}{@{}p{2.6cm} p{3.8cm} p{5.5cm}@{}}
\toprule
\textbf{Skill} & \textbf{Source} & \textbf{Description} \\
\midrule
\multicolumn{3}{l}{\emph{\textsc{SpreadsheetBench} Verified}: spreadsheet and data workflows} \\
\midrule
\rowcolor{tab-row-even} xlsx & fcakyon/claude-codex-settings & Read, edit, create, and convert spreadsheet files (.xlsx, .xlsm, .csv, .tsv) with formula-aware workflows. \\
spreadsheet & fcakyon/claude-codex-settings & Spreadsheet creation, editing, analysis, and formatting with cached recalculation and visual review. \\
\rowcolor{tab-row-even} xlsx-toolkit & borghei/claude-skills & Audit Excel workbooks for sheet count, formula density, external references, named ranges, and data validation. \\
data-analysis & claude-office-skills/skills & Analyze spreadsheet data, generate insights, create visualizations, and build reports from Excel/CSV data. \\
\rowcolor{tab-row-even} chart-designer & claude-office-skills/skills & Design data visualizations and chart configurations for ECharts, Chart.js, and other charting libraries. \\
\midrule
\multicolumn{3}{l}{\emph{\textsc{WebArena} Verified}: browser automation and web interaction} \\
\midrule
browser-automation & web-infra-dev/midscene-skills & Vision-driven browser automation using Midscene; operates from screenshots without DOM access. \\
\rowcolor{tab-row-even} vitest-midscene-e2e & web-infra-dev/midscene-skills & AI-powered E2E testing across web, Android, and iOS using natural-language UI interactions. \\
browsing-with-playwright & mjunaidca/mjs-agent-skills & Browser automation via Playwright MCP for navigation, form filling, screenshots, and data extraction. \\
\rowcolor{tab-row-even} agent-browser & fcakyon/claude-codex-settings & Snapshot-and-ref browser workflow covering navigation, element interaction, tab management, and auth. \\
webapp-testing & ComposioHQ/awesome-claude-skills & Toolkit for testing local web applications using Playwright with screenshot capture and log viewing. \\
\midrule
\multicolumn{3}{l}{\emph{\textsc{SWE-bench} Verified}: software engineering practices} \\
\midrule
\rowcolor{tab-row-even} test-driven-development & obra/superpowers & Write failing tests before implementation; red--green--refactor cycle for features and bugfixes. \\
python-testing-patterns & wshobson/agents & Comprehensive pytest strategies with fixtures, mocking, parameterization, and test-driven development. \\
\rowcolor{tab-row-even} refactor & github/awesome-copilot & Surgical code refactoring: extract functions, rename variables, break down god functions, improve type safety. \\
security-review & affaan-m/everything-claude-code & Security checklist for authentication, user input handling, secrets, API endpoints, and payment features. \\
\rowcolor{tab-row-even} clean-code & sickn33/antigravity-awesome-skills & Principles from ``Clean Code'' by Robert C.\ Martin; transform working code into clean, maintainable code. \\
\bottomrule
\end{tabular}
\end{table}

All skills are available at \texttt{https://skills.sh/\{source\}} where \texttt{\{source\}} is the repository path listed in the Source column above.

\subsection{\textsc{HumanEval+} Synthetic Skill Construction}
\label{app:humaneval_seeds}

\begingroup\widowpenalties 2 10000 10000
For the controlled experiments on HumanEval+, we construct synthetic skills using a fixed seed pool of 12 coding-practice directives. Each directive is fed as a creation prompt to the official \texttt{skill-creator} tool, which generates a complete \texttt{SKILL.md} file containing typed behavioral constraints, scope annotations, and metadata. By sampling subsets of 1--6 seeds per configuration, we control the number of skills and the total constraint count while preserving the naturalness of machine-generated skill text.\par
\endgroup

The 12 seed directives are listed in Table~\ref{tab:humaneval_seeds}. They encode contrasting programming preferences: \emph{Functional Purity} emphasizes pure functions and immutable data, while \emph{OOP Encapsulation} favors class-based organization; \emph{Maximal Conciseness} favors compact expressions, while \emph{Defensive Validation} emphasizes explicit checks; and \emph{Performance First} and \emph{Algorithmic Elegance} favor iterative and recursive formulations, respectively. Whether a pair forms a hard conflict, a soft conflict, or remains compatible depends on the generated directives and their activation scopes. The synthetic audit evaluates 60 labeled incompatible pairs separately from Preservation over 60 source constraints.

\begin{table}[!ht]
\centering
\caption{Seed pool for \textsc{HumanEval+} synthetic skill construction. Each seed is a creation prompt fed to \texttt{skill-creator}. Intended contrasting seed pairs are indicated in parentheses.}
\label{tab:humaneval_seeds}
\footnotesize
\setlength{\tabcolsep}{4pt}
\renewcommand{\arraystretch}{1.15}
\begin{tabular}{@{}c p{2.8cm} p{8.5cm}@{}}
\toprule
\textbf{\#} & \textbf{Seed Name} & \textbf{Creation Prompt (abbreviated)} \\
\midrule
\rowcolor{tab-row-even} 1 & Type Safety & Enforce comprehensive type annotations on all parameters and returns; enable strict mypy-level checking; add runtime type guards for public APIs. \\
2 & Functional Purity & Write pure functions with no side effects; prefer immutable data structures; use function composition over imperative control flow. \emph{(contrast: \#10)} \\
\rowcolor{tab-row-even} 3 & Defensive Validation & Validate all inputs at function boundaries; check boundary conditions explicitly; raise informative errors for degenerate cases. \emph{(contrast: \#6, \#11)} \\
4 & Performance First & Prefer iterative over recursive solutions; use generators for memory efficiency; avoid unnecessary object allocations. \emph{(contrast: \#12)} \\
\rowcolor{tab-row-even} 5 & Documentation \& Readability & Require NumPy-style docstrings; use descriptive variable names ($\ge$3 chars); add inline comments for non-obvious logic. \emph{(contrast: \#6)} \\
6 & Maximal Conciseness & Favor single-expression solutions; use list comprehensions and ternary operators; minimize line count while preserving correctness. \emph{(contrast: \#3, \#5)} \\
\rowcolor{tab-row-even} 7 & Robust Error Handling & Use custom exception classes; never use bare \texttt{except}; ensure resource cleanup with context managers; provide traceable error messages. \\
8 & Security Awareness & Sanitize all string inputs; avoid \texttt{eval}/\texttt{exec}; use safe formatting; guard against injection patterns in string-processing tasks. \\
\rowcolor{tab-row-even} 9 & Test-First & Write assertion-based test cases before implementation; cover edge cases; validate output types and boundary values. \\
10 & OOP Encapsulation & Use classes with private attributes; follow single-responsibility principle; prefer composition over inheritance; implement \texttt{\_\_repr\_\_} and \texttt{\_\_eq\_\_}. \emph{(contrast: \#2)} \\
\rowcolor{tab-row-even} 11 & Pythonic Idioms & Leverage built-in functions (\texttt{zip}, \texttt{enumerate}, \texttt{any/all}); use unpacking and walrus operator; avoid C-style index loops. \emph{(contrast: \#3)} \\
12 & Algorithmic Elegance & Prefer recursive divide-and-conquer formulations; document time/space complexity; implement solutions at optimal asymptotic bounds. \emph{(contrast: \#4)} \\
\bottomrule
\end{tabular}
\end{table}

For the scaling analysis (Section~\ref{sec:scaling}), the constraint-count axis fixes a random 4-seed subset and varies the total constraints by controlling the \texttt{skill-creator}'s output granularity parameter (20--140 constraints total). The skill-count axis fixes 20 constraints per skill and samples 1--6 seeds without replacement. Each configuration is evaluated over all 164 HumanEval+ tasks with 20 runs; the reported intervals and record-sharing protocol appear in Appendix~\ref{app:new_experiments}.

\end{document}